\documentclass[runningheads]{llncs}
\usepackage{graphicx}
\usepackage{amsmath,amssymb} 
\usepackage{color}

\usepackage{tabularx}
\usepackage{multirow}
\usepackage{multicol}
\usepackage[scriptsize]{subfigure}
\usepackage{caption}
\usepackage{booktabs}  

\usepackage{pifont}
\newcommand{\cmark}{\ding{51}}%
\newcommand{\xmark}{\ding{55}}%

\usepackage[normalem]{ulem}
\useunder{ine}{}{}

\def\ie{\emph{i.e.}}
\def\eg{\emph{e.g.}}

\def\etal{\emph{et al.}}

\usepackage{array}
\newcolumntype{H}{>{\setbox0=\hbox\bgroup}c<{\egroup}@{}}

\usepackage[normalem]{ulem}
\useunder{\uline}{\ul}{}

\usepackage{appendix}

\usepackage[export]{adjustbox}

\begin{document}
\pagestyle{headings}
\mainmatter


\title{CVLNet: Cross-View Semantic Correspondence Learning for Video-based Camera Localization} 
\titlerunning{ }
\authorrunning{ }


\author{Yujiao Shi\textsuperscript{\rm 1},~~
Xin Yu\textsuperscript{\rm 2},~~
Shan Wang\textsuperscript{\rm 1},~~
Hongdong Li\textsuperscript{\rm 1}}

\institute{\textsuperscript{\rm 1}Australian National University~~
\textsuperscript{\rm 2}University of Technology Sydney\\
}

\maketitle

\begin{abstract}
This paper tackles the problem of Cross-view Video-based camera Localization (CVL). 
The task is to localize a query camera by leveraging information from its past observations, \ie, a continuous sequence of images observed at previous time stamps, and matching them to a large overhead-view satellite image. 
The critical challenge of this task is to learn a powerful global feature descriptor for the sequential ground-view images while considering its domain alignment with reference satellite images. 
For this purpose, we introduce CVLNet, which first projects the sequential ground-view images into an overhead view by exploring the ground-and-overhead geometric correspondences and then leverages the photo consistency among the projected images to form a global representation. 
In this way, the cross-view domain differences are bridged. 
Since the reference satellite images are usually pre-cropped and regularly sampled, there is always a misalignment between the query camera location and its matching satellite image center. 
Motivated by this, we propose estimating the query camera's relative displacement to a satellite image before similarity matching. 
In this displacement estimation process, we also consider the uncertainty of the camera location. For example, a camera is unlikely to be on top of trees. 
To evaluate the performance of the proposed method, we collect satellite images from Google Map for the KITTI dataset and construct a new cross-view video-based localization benchmark dataset, KITTI-CVL. 
Extensive experiments have demonstrated the effectiveness of video-based localization over single image-based localization and the superiority of each proposed module over other alternatives.

\end{abstract}


\section{Introduction}
\label{sec:intro}

Cross-view image-based localization using ground-to-satellite image matching has attracted significant attention these days~\cite{vo2016localizing,Hu_2018_CVPR,Liu_2019_CVPR,Regmi_2019_ICCV,Cai_2019_ICCV,shi2019spatial,shi2020optimal,shi2020looking,zhu2021revisiting,toker2021coming,zhu2021vigor}.
It has found many practical applications such as autonomous driving and robot navigation.
Prior works have been focused on localizing omnidirectional ground-view images with a $360^\circ$ Field-of-View (FoV), which helps to provide rich and discriminative features for localization. 
However, when a regular forward-looking camera with a limited FoV is used, those omnidirectional camera-based algorithms suffer severe performance degradation.

\begin{figure}[t!]
    \centering
    \subfigure[\scriptsize Cross-view image-based localization]{
    \centering
    \parbox[][2.5cm][c]{0.48\linewidth}{
    \includegraphics[width=\linewidth]{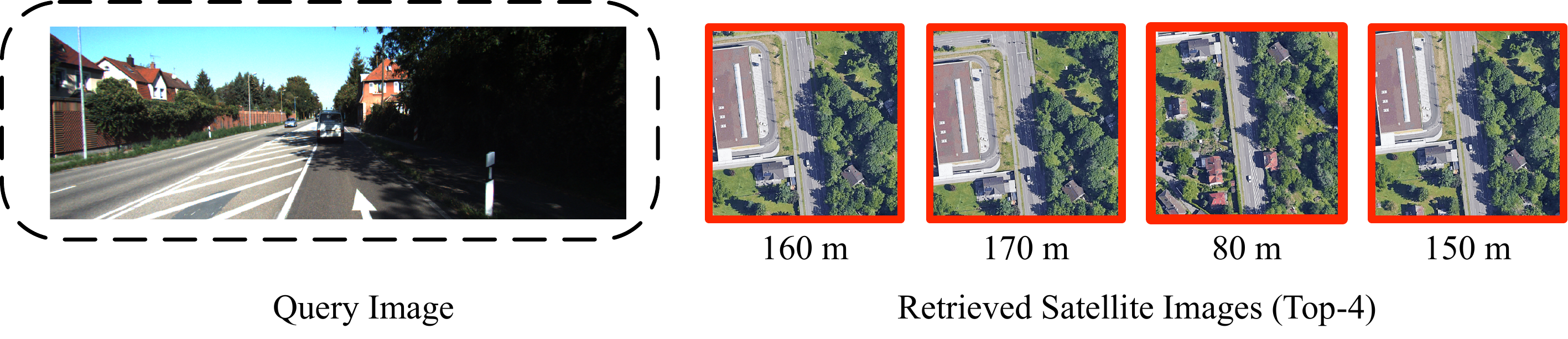}}}
    \subfigure[\scriptsize Cross-view video-based localization]{
    \centering
    \parbox[][2.5cm][c]{0.48\linewidth}{
    \includegraphics[width=\linewidth]{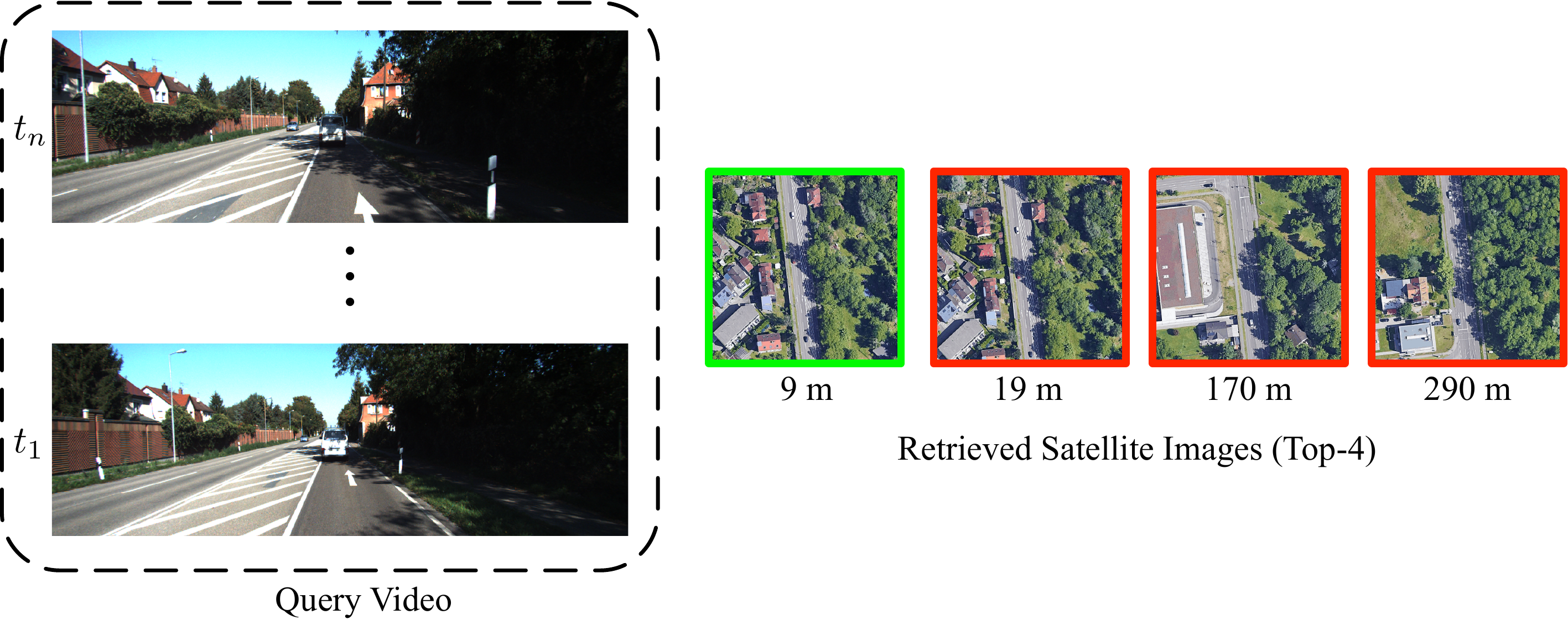}}}
    \caption{\scriptsize 
    Single-frame image-based localization (a) Vs. Multi-frame video-based localization (b). 
    The multi-frame video-based localization leverages richer scene context of a query place, increasing the discriminating power of query descriptors compared to single image-based localization. 
    As a result, the matching satellite image, marked by green border, from the database is more likely to be retrieved. 
    Red border indicates non-matching satellite images to the query image. 
    }
    \label{fig:intro} 
\end{figure}

To tackle this challenge, this paper proposes to use a continuous short video, \ie, a sequence of ground-view images, as input for the task of visual localization. 
Specifically, we localize a camera at the current time stamp $t_n$ by augmenting it with previous observations at time \ie, $t_{1}\sim t_{n-1}$, as shown in Fig.~\ref{fig:intro}. 
Compared to using a single query image, a short video provides richer visual and dynamic information about the current location. 

We present a Cross-view Video-based Localization Network, named CVLNet, to address the camera localization problem. To the best of our knowledge, our CVLNet is the first vision- and deep-based cross-view geo-localization framework that exploits a continuous video rather than a single image to pinpoint the camera location.

Our CVLNet is composed of two branches that extract deep features from ground and satellite images, respectively.
Considering the drastic viewpoint changes between the two-view images, we first introduce a Geometry-driven View Projection (GVP) module to transform ground-view features to the overhead view by explicitly exploring their geometric correspondences. 
Then, we design a Photo-consistency Constrained Sequence Fusion (PCSF) module to fuse the sequential features. 
Our PCSF first estimates the reliability of the sequential ground-view features in overhead view by leveraging photo-consistency across them and then aggregates them as a global query descriptor. 
In this manner, we achieve more discriminative and reliable ground-view feature representation. 

Since satellite images in a database are usually pre-cropped and sampled at discretized locations, there would be a misalignment between a query camera location and its matching satellite image center. 
Furthermore, a query camera is usually impossible in some regions (\eg, on top of a tree), while likely on the other areas (\eg, road). 
Hence, we propose a Scene-prior driven Similarity Matching (SSM) strategy to estimate the relative displacement between a query camera location and a satellite image center while restricting the search space by scene priors. 
The scene priors are learned statistically from training rather than pre-defined. 
With the help of SSM, our CVLNet can eliminate unreasonable localization results.


In order to train and evaluate our method, we curate a new cross-view dataset by collecting satellite images for the KITTI dataset~\cite{geiger2013vision} from Google Map~\cite{google}.
The new dataset combines sequential ground-view images from the original KITTI dataset and the newly collected satellite images. 
To the best of our knowledge, it is not only the first cross-view video-based localization dataset, but also the first cross-view localization dataset where ground-view images are captured by a perspective pin-hole camera with a restricted FoV (rather than being cropped from Google street-view panoramas~\cite{vo2016localizing,shi2020looking}). Extensive experiments on the newly collected dataset demonstrate that our method effectively localizes camera positions and outperforms the state-of-the-art remarkably.


\section{Related Work}

\textbf{Image-based localization.}
The image-based localization problem is initially tackled as a ground-to-ground image matching~\cite{arandjelovic2016netvlad,kim2017learned,liu2019stochastic,noh2017large,ge2020self,zhou2020da4ad}, where both the query and database images are captured at the ground level. 
However, those methods cannot localize query images when there is no corresponding reference image in the database. 
Thanks to the wide-spread coverage and easy accessibility of satellite imagery, recent works~\cite{castaldo2015semantic,lin2013cross,mousavian2016semantic,vo2016localizing,tian2017cross,Hu_2018_CVPR,Liu_2019_CVPR,Regmi_2019_ICCV,Cai_2019_ICCV,shi2019spatial,hu2020image,shi2020optimal,shi2020looking,zhu2021revisiting,toker2021coming,zhu2021vigor,shi2022accurate,Zhu_2022_CVPR,elhashash2022cross,guo2022soft,zhao2022mutual} resort to satellite images for city-scale localization. 

While recent works on city-scale ground-to-satellite localization have achieved promising results, they mostly focus on localizing isolated omnidirectional ground images. 
When the query camera has a limited FoV, we propose using a continuous video instead of a single image for camera localization, improving the discriminativeness of the query location representation.

\textbf{Video-based localization. } 
The concept of video-based localization can be divided into three main categories; Visual Odometry (VO) ~\cite{bloesch2015robust,leutenegger2015keyframe,chien2016WhenTU}, Visual-SLAM (vSLAM) ~\cite{cadena2016past,engel2014lsd,klein2007parallel,mur2015orb,mur2017orb,campos2021orb} and Visual Localization ~\cite{mur2017visual,wolcott2014VisualLW,voodarla2021sbev,stenborg2018long,stenborg2020using,vaca2012city,regmi2021video}. VO techniques can be classified according to their camera setup — either monocular or stereoscopic
or their processing techniques — either feature-based or appearance-based. VO methods usually use a combination of feature tracking and feature matching ~\cite{Yousif2015AnOT,Scaramuzza2011VisualO}. vSLAM pertains to simultaneously creating a map of features and localizing the robot in that map, all using visual information ~\cite{gao2018ldso,kasyanov2017keyframe}. Many a time, the map is pre-built, and the robot needs to localize itself using camera-based map-matching, which is referred to as Visual Localization ~\cite{liu2020visual}. Even though these methods use a series of image frames to determine the robot's location, they match information from the same viewpoint. In our work, we have developed a cross-view video-based localization approach by leveraging a sequence of images with varied viewpoints and limited FoVs, aiming to improve the representativeness of a query location significantly.

 \begin{figure*}[ht!]
 \setlength{\abovecaptionskip}{0pt}
    \setlength{\belowcaptionskip}{0pt}
    \centering
    \includegraphics[width=0.8\linewidth]{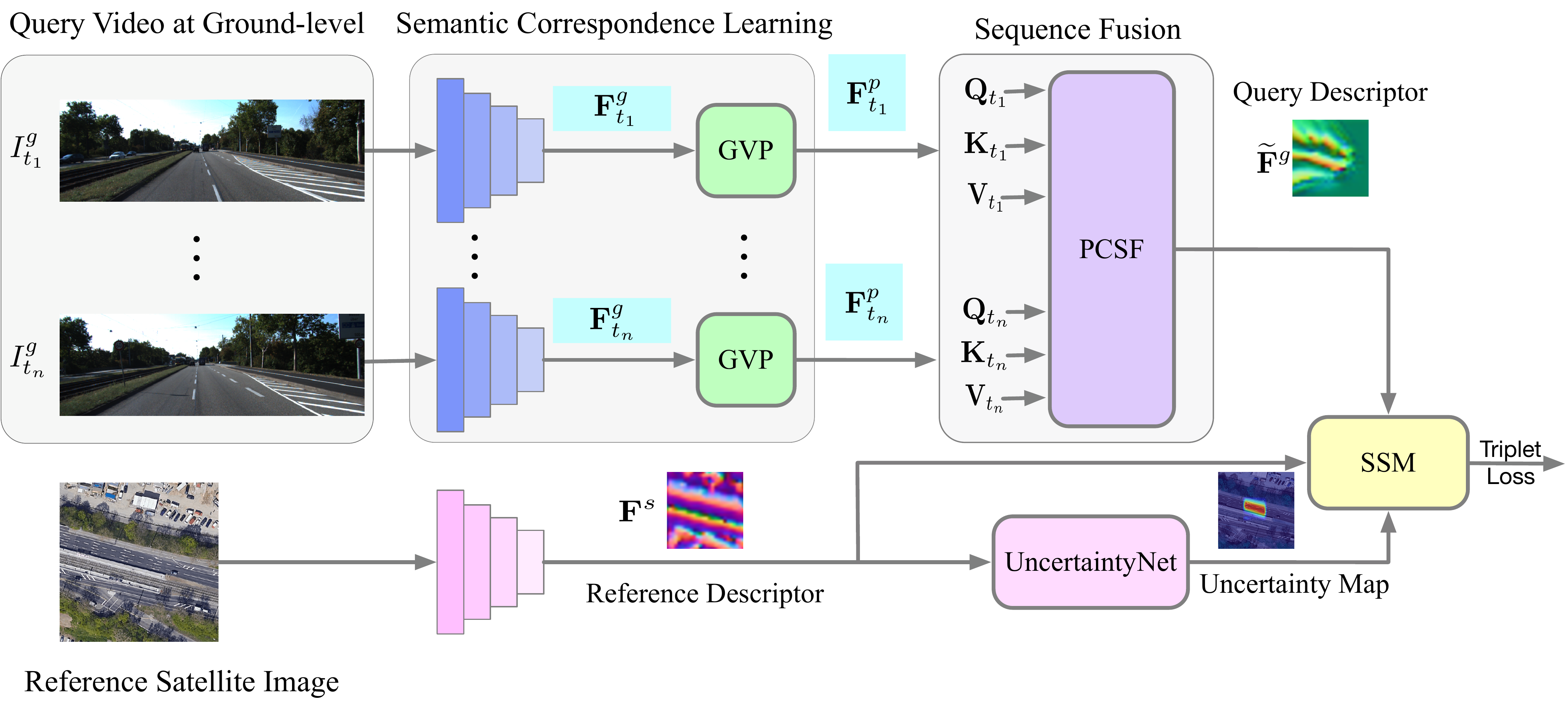}
    \caption{\scriptsize Overview of our proposed CVLNet. Our Geometry-driven View Projection (GVP) module first aligns the sequential ground-view features in the overhead view and presents them in a unified coordinate system. Next, the Photo-consistency Constrained Sequential Fusion (PCSF) module measures the photo-consistency of an overhead view pixel across the different ground-views and fuses them together, obtaining a global feature representation $\widetilde{\mathbf{F}}^g$ of the query video. The global feature representation is then compared with the satellite feature map $\mathbf{F}^s$ with a Scene-prior driven Similarity Matching (SSM) scheme to determine the relative displacement between the query camera location and the satellite image center, guided by an uncertainty map. After alignment, the feature similarity is then computed for image retrieval.}
    \label{fig:framework_CVL}
\end{figure*}

\section{CVLNet: Cross-view Video-based Localization}
This paper tackles the ground-to-satellite localization task. 
Instead of using a single query image captured at the ground level, we augment the query image with a short video containing previous observations. 
To solve this task, our motivation is first projecting the images in the ground video to an overhead\footnote{For clarity, we use ``overhead'' throughout the paper to denote the projected features from ground-views, and ``satellite'' to indicate the real satellite image/features.} perspective and then extracting a global description from the projected image sequence for localization. 
An overview of our pipeline is illustrated in Fig.~\ref{fig:framework_CVL}.

\subsection{Geometry-driven view projection (GVP)} 
Prior methods often resort to a satellite to ground projection to bridge the cross-view domain gap. This is achieved either by a polar transform~\cite{shi2019spatial,shi2020looking,toker2021coming} or a projective transform~\cite{shi2022geometry,shi2022accurate,shi2022beyond}. 
However, both transforms need to know the query camera location with respect to the satellite image center. 
In the CVUSA and CVACT dataset where polar transform performs excellent, the query images accidentally align with their matching satellite image center, which however does not occur in practice. 
When there is a large offset between the real camera location and its assumed location with respect to its matching satellite image (\eg, satellite image center in polar transform), the performance will be impeded significntly. 
Hence, instead of projecting satellite images to ground views, we introduce a Geometry-driven View Projection (GVP) module to transform ground-view images to overhead view.

\begin{figure}[t!]
    \setlength{\abovecaptionskip}{0pt}
    \setlength{\belowcaptionskip}{0pt}
    \centering
    \begin{minipage}{0.48\linewidth}
    \includegraphics[width=0.85\linewidth]{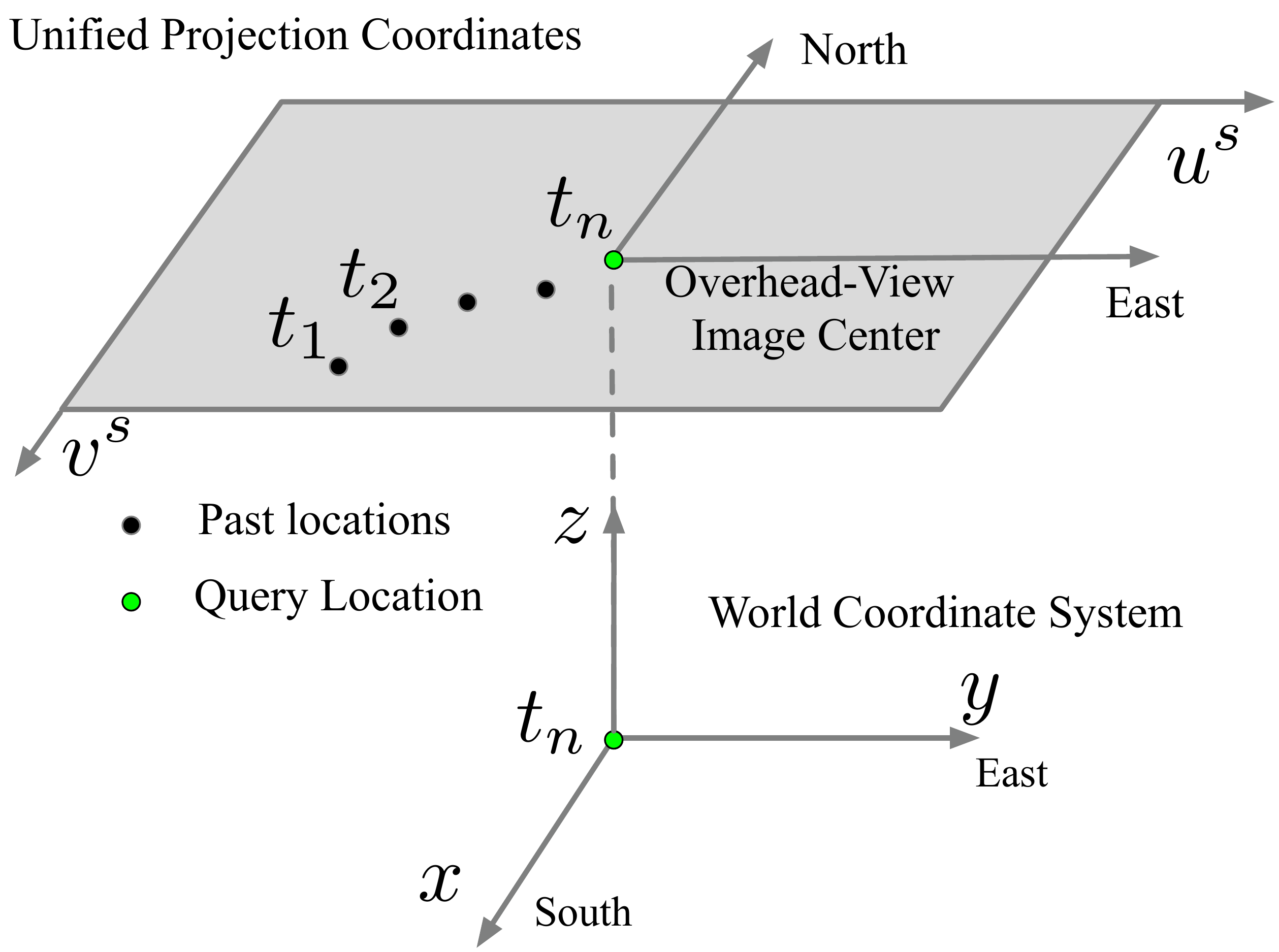}
    \captionof{figure}{ 
    A unified projection coordinates for the projected overhead-view features. 
    Ground-view observations at timestamps $T=\{t_1,t_2,\dots,t_n\}$ are projected to the same overhead-view grid with the center corresponding to the query camera location at $t_n$.  }
    \label{fig:projection}
    \end{minipage}
    \hfill
    \begin{minipage}{0.48\linewidth}
    \includegraphics[width=\linewidth]{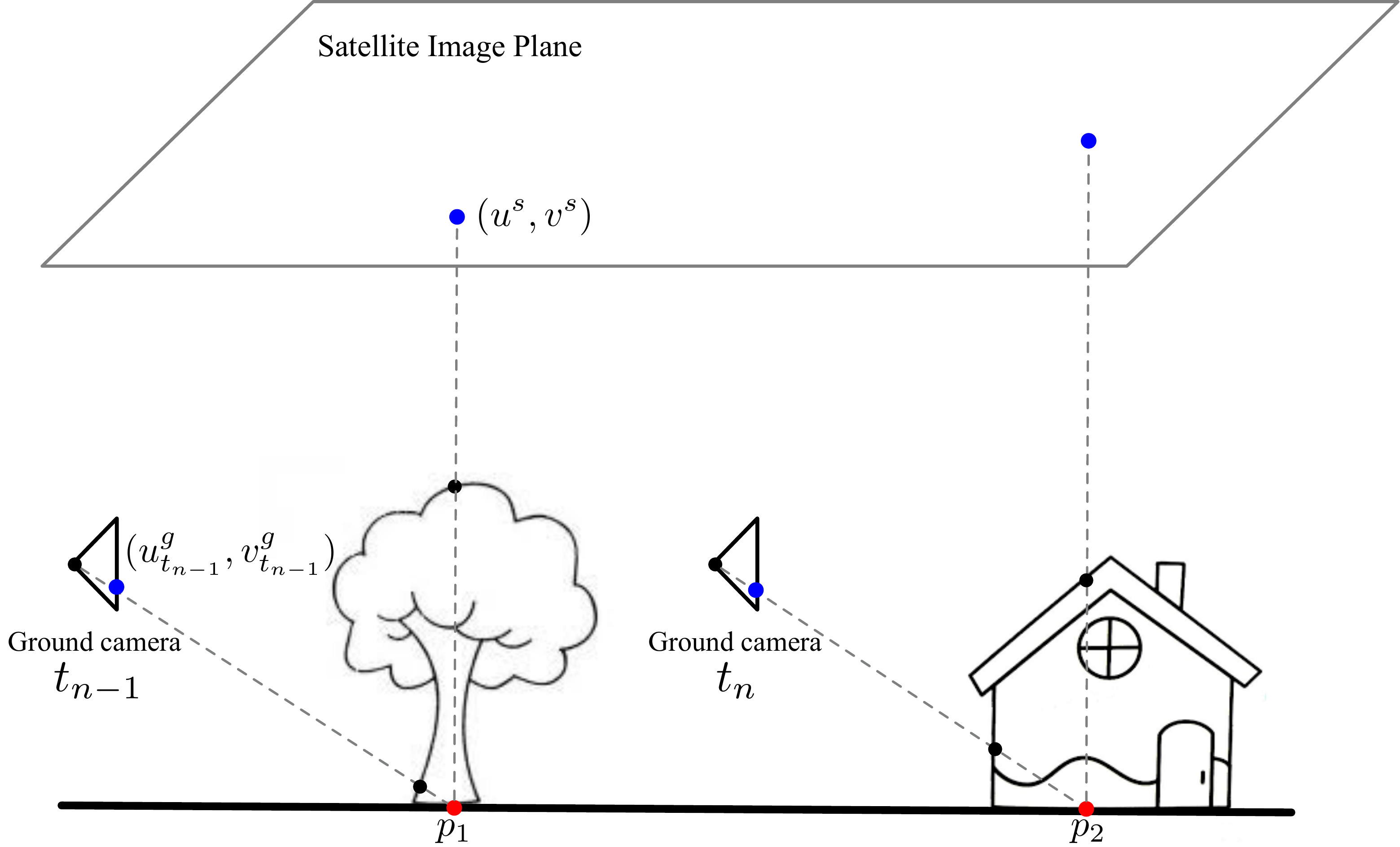}
    \captionof{figure}{Geometry-driven cross-view semantic correspondence learning. 
    Both tree canopy and tree trunk are ``trees''. Building roof and facades are both ``buildings".}
    \label{fig:geometry}
    \end{minipage}
\end{figure}

Starting from a blank canvas in the overhead view with its center corresponds to the geospatial location of the query camera, we aim to fill it with features collected from ground-view images. 
We set the origin of the world coordinate system to the geo-spatial query camera location as well, with its $x$-axis pointing to the south direction, $y$ axis pointing to the east direction, and the $z$-axis vertically upward.
Different ground-view images in a video sequence are projected to the same overhead-view coordinate system so that they are geographically aligned after projection. 
Fig.~\ref{fig:projection} provides a visual illustration of the coordinate systems.

\noindent\textbf{Parallel projection of a satellite camera.}
The projection between the satellite image coordinate system $(u^s, v^s)$ and the world coordinate system $(x, y, z)$ can be approximated as a parallel projection~\cite{shi2022geometry}, 
  $[x, y]^T = \lambda [v^s - v^s_0, u^s - u^s_0]^T$,
where $(u_0^s, v_0^s)$ indicates the satellite map center, $\lambda$ indicates the real-world distance between two neighboring pixels in the satellite map. 

\noindent\textbf{Perspective projection of ground-view images.}
Denote $\mathbf{R}_{t_i}$ and $\mathbf{t}_{t_i}$ as the rotation and translation for the camera at time step $t_i$ in the world coordinate, $\mathbf{E}_{t_i}$ as the camera intrinsic, and $N$ as the sequence number. 
The relative $\mathbf{R}_{t_i}$ and $\mathbf{t}_{t_i}$ can be easily obtained by Structure from Motion~\cite{schonberger2016structure}. 
The projection between the world coordinate system $(x, y, z)$ and the ground-view camera coordinate system $ (u^g_{t_i}, v^g_{t_i})$ is expressed as
    $w_{t_i} [u^g_{t_i},  v^g_{t_i},  1]^T = \mathbf{E}_{t_i} [\mathbf{R}_{t_i},  \mathbf{t}_{t_i}] [x,  y,  z,  1]^T$,
where $w_{t_i}$ is a scale factor in the perspective projection.

\noindent\textbf{Ground-to-satellite projection.} 
There is a height ambiguity of satellite pixels in the ground-to-satellite projection. 
Instead of explicitly estimating the heights, we present a simple yet effective solution. 
Specifically, we project ground-view observations to the overhead view assuming satellite pixels lie on the ground plane. 
Rather than projecting original image RGB pixels, we project high-level deep features. 
The geometric projection from the ground-view to the overhead-view is derived as,
\begin{equation}
\small
 w_{t_i} [ u^g_{t_i}, v^g_{t_i}, 1]^T = \mathbf{E}_{t_i} [\mathbf{R}_{t_i},  \mathbf{t}_{t_i}] [\lambda( v^s - v^s_0), \lambda(u^s - u^s_0), -h, 1]^T. 
\label{eq:g2s}
\end{equation}where $h$ is the height of the query camera with respect to the ground plane, and $w_{t_i}$ can be computed from the above equation.

Denote $\mathbf{F}_{t_i}^g \in \mathbb{R}^{H\times W \times C} $ as ground-view image features by a CNN backbone, where $H$, $W$ and $C$ are the height, width and channels of the features, respectively, and $\text{GVP}(\cdot)$ as the geometry-driven view projection operation illustrated in Eq.~\eqref{eq:g2s}. 
The projected features in overhead view are then obtained by 
$\mathbf{F}_{t_i}^p = \text{GVP}(\mathbf{F}_{t_i}^g),~~\mathbf{F}_{t_i}^p \in \mathbb{R}^{S\times S \times C}$,
where $S$ indicates the overhead-view feature map resolution. 

This projection establishes the exact geometric correspondences between the ground and overhead views for scene contents on the ground plane. 
For scene objects with higher heights, projecting features rather than image pixels can alleviate the strict constraint while providing a cue that corresponding objects exist between the views. 
As shown in Fig.~\ref{fig:geometry}, for pixel $(u^s, v^s)$, the projected feature from the ground-view at $t_{n-1}$ represents the tree trunk, but the feature in the satellite image corresponds to the tree canopy.  
Both tree canopy and tree trunk indicate there is a tree at location $p_2$. 
Then, by applying a matching loss between the two features (tree trunk and tree canopy), the network will be trained to learn viewpoint invariant features, \ie, both tree trunk and tree canopy are mapped to the semantic features of ``tree''. 

The coverage of the canvas for ground-to-satellite projection is set to the reference satellite image coverage, \ie, around $100\text{m} \times 100\text{m}$, with its center corresponding to the query camera location. When the sequence is too long with some previous image contents exceeding the canvas's pre-set coverage, the exceeded contents will not be collected. This is because scene contents that are too far from the query camera location are less important for localization, and it is better to cover most of the synthetic overhead-view feature map by referencing satellite images.

\subsection{Photo-consistency constrained sequence fusion}
\label{sec:seq_fusion}
We leverage photo consistency among different ground-views for the video sequence fusion. 
For a satellite pixel, when its corresponding features in several (more than two) ground views are similar, 
the existence of a scene object at this geographical location is highly reliable for these ground-views. 
We should highlight these corresponding features when generating descriptors for scene contents.  
Driven by this, we design a Photo-consistency Constrained Sequence Fusion (PCSF) module. 
Our PCSF module employs an attention mechanism~\cite{vaswani2017attention} to emphasize reliable features in fusing a video sequence and obtaining a global descriptor for the video.

Our GVP block has aligned the original ground-view features at different time steps in a unified overhead-view coordinate. 
When the features of a geographical location observed by different ground views are similar, those features should be more reliable for localization. 
We leverage the self-attention mechanism~\cite{vaswani2017attention} to measure the photo-consistency/similarity across different views and find reliable features.  
Specifically, for each projected feature map $\mathbf{F}^p_{t_i}$ at time step $t_i$, we compute its query, key and value by two stacked convolutional layers, denoted by $\mathbf{Q}_{t_i}, \mathbf{K}_{t_i}, \mathbf{V}_{t_i} \in \mathbb{R}^{S \times S \times C}$, respectively. 
The stacked convolutional layers increase the receptive field and the representative ability of the key, query, and value features at each spatial location. 
Next, we compute the similarities between each projected feature map at $t_i$ and other projected feature maps at $t_j$, $i, j = 1, ..., N$, and normalize them across all possible $j$ by a softmax operation, expressed as,

\begin{equation}
    \mathbf{M}_{i, j} = \text{Softmax}_j \left ( \mathbf{Q}_{t_i}^T \mathbf{K}_{t_j} \right ), \quad \mathbf{M} \in \mathbb{R}^{N \times N \times S \times S}.
    \label{eq:similarity}
\end{equation}
The final fused feature is obtained by,
\begin{equation}
    \widetilde{\mathbf{F}}^g = \frac{1}{N} \sum\nolimits_i^N \sum\nolimits_j^N \mathbf{M}_{i, j} \mathbf{V}_{t_j}, \quad \widetilde{\mathbf{F}}^g \in \mathbb{R}^{S \times S \times C}.
    \label{eq:aggregation}
\end{equation}
In this way, we highlight the common features across the views and make the global descriptor reliable.

\subsection{Scene-prior driven similarity matching}
\label{sec:uncertainty}


We want to address the location misalignment between a query camera location and its matching satellite image center by ground-to-satellite projection and spatial correlation between the projected features and the real satellite features. 
Hence, the satellite feature descriptors should be translational equivariant, which is an inherent property of conventional CNNs. 
Following most previous works~\cite{Hu_2018_CVPR,shi2019spatial,shi2020optimal,shi2020looking,zhu2021vigor}, we use VGG16~\cite{Simonyan2014VeryDC} as our backbone for satellite (and ground) feature extraction. 
The extracted satellite features, denoted as $\mathbf{F}^s \in \mathbb{R}^{S \times S \times C}$, share the same spatial scale as the global representation of the query video. 
Next, we adopt a Normalized spatial Cross-Correlation (NCC) to estimate latent alignment between the query location and a satellite image center.

Denote $[\mathbf{F}^s]_{m, n}$ as a shifted version of a satellite feature map with its center at $(m, n)$ in the original satellite feature map, and $m=0, n=0$ correspond to the center of the original satellite feature map. 
The similarity between $\mathbf{F}^s$ and $\widetilde{\mathbf{F}}^g$ aligned at $(m, n)$ computed by NCC is,
\begin{equation}
    \mathbf{D}_0(\mathbf{F}^s, \widetilde{\mathbf{F}}^g)_{m, n} =  \frac{[\mathbf{F}^s]_{m, n} \cdot \widetilde{\mathbf{F}}^g}{\| [\mathbf{F}^s]_{m, n} \|_2 \| \widetilde{\mathbf{F}}^g \|_2}, 
    \label{eq:ncc}
\end{equation}
where $\mathbf{D}_0(\mathbf{F}^s, \widetilde{\mathbf{F}}^g) \in \mathbb{R}^{h \times w}$ denotes the similarity matrix between $\mathbf{F}^s$ and $\widetilde{\mathbf{F}}^g$ at all possible spatial-aligned locations, $m \in [-\frac{h}{2}, \frac{h}{2}]$, and $n \in [-\frac{w}{2}, \frac{w}{2}]$. 
A potential spatial-aligned location of the satellite map lies in a region of $10 \times 10 \ \text{m}^2$ in our KITTI-CVL dataset, as the database satellite image is collected very ten meters.

To exclude impossible query camera locations, \eg, top of trees, we estimate an uncertainty map from the satellite semantic features, 
    $\mathbf{U}(\mathbf{F}^s) = \mathcal{U}(\mathbf{F}^s)$, 
    $\mathbf{U}(\mathbf{F}^s)\in \mathbb{R}^{h \times w}$,  
where $\mathcal{U}(\cdot)$ is the uncertainty net, composed of a set of convolutional layers.
The value of each element in $\mathbf{U}(\mathbf{F}^s)$ is within the range of $[0, 1]$, forced by a Sigmoid layer. 
By encoding the uncertainty, The similarity between $\mathbf{F}^s$ and $\widetilde{\mathbf{F}}^g$ aligned at $(m, n)$ is then written as,
\begin{equation}
    \mathbf{D}(\mathbf{F}^s, \widetilde{\mathbf{F}}^g)_{m, n} =\frac{\mathbf{D}_0(\mathbf{F}^s, \widetilde{\mathbf{F}}^g)_{m, n}}{\mathbf{U}(\mathbf{F}^s)_{m, n}}.
    \label{eq:ncc_uncertainty}
\end{equation}
When the uncertainty at $(m, n)$ is large, the similarity between $\mathbf{F}^s$ and $\widetilde{\mathbf{F}}^g$ aligned at this location will be decreased. 
We do not have explicit supervisions for the uncertainty map. Rather, it is learned statistically from training. 
The relative displacement between $\mathbf{F}^s$ and $\widetilde{\mathbf{F}}^g$ is obtained by,
\begin{equation}
    m^*, n^* = \mathop{
    \arg\max}_{m, n}\mathbf{D}(\mathbf{F}^s, \widetilde{\mathbf{F}}^g)_{m, n}.
\end{equation}

During inference, we have no idea which one is the matching reference image for a query image. Thus the uncertainty-guided similarity matching is applied to all reference features (including non-matching ones). 
Furthermore, it is more challenging when a similarity score between non-matching ground and satellite features is high. 
Hence, we apply the similarity matching scheme to the pairs of query and non-matching reference images as well during training and minimize their maximum similarity, making the learned features more discriminative.

\subsection{Training objective}
We employ the soft-weighted triplet loss~\cite{Hu_2018_CVPR} to train our network. 
The loss includes a positive term to maximize the similarity between the matching query and reference pairs and a negative term to minimize the similarity between non-matching pairs. 
The non-matching term also prevents our view projection module from trivial solutions. Therefore, it is formulated as,
\begin{equation}
    \small
    \mathcal{L} = \log \left ( 1 + e^{\alpha \left(d(\widetilde{\mathbf{F}}^g, \mathbf{F}^s) - d(\widetilde{\mathbf{F}}^g, \mathbf{F}^{s^{*}})\right)} \right ), 
    \label{eq:loss}
\end{equation}
where $\mathbf{F}^s$ is the matching satellite image feature to the ground feature $\mathbf{F}^g$, $\mathbf{F}^{s^{*}}$ is the non-matching satellite image feature, $d(\cdot, \cdot)$ is the ${L}_2$ distance between its two inputs after alignment, and $\alpha$ is set to 10. 


\section{The KITTI-CVL Dataset}
KITTI is one of the widely used benchmark datasets for testing computer vision algorithms for autonomous driving ~\cite{geiger2013vision}.  In this paper, we intend to investigate a method for using a short video sequence for satellite image-based camera localization.
For this purpose, we supplement the KITTI drive sequences with corresponding satellite images.  This is done by cropping high-definition Google earth satellite images using the KITTI-provided GPS tags for vehicle trajectories.  Based on these GPS tags of the ground-view images, we select a large region that covers the vehicle trajectory.  We then uniformly partition the region into overlapping satellite image patches. Each satellite image patch has a resolution of $1280\times 1280$ pixels, amounting to about 20 cm per pixel.

\smallskip
\noindent\textbf{Training, Validation and Test sets. }
The KITTI data contains different trajectories captured at different time. 
In our Training, Validation and Test set split, the images of Training and Validation set are from the same region.
The Validation set is constructed in this way to select the best model during training. 
In contrast, the images in the test set are captured at different regions from the Training and Validation sets.
The test set aims to evaluate the generalization ability of the compared algorithms.  

Only the nearest satellite image for each ground image in the sampled grids is retained for the Training and Validation set. 
We use the same method to construct our first test set, Test-1. 
Furthermore, we construct the second test set, Test-2, where all satellite images in the sampled grids are reserved. 
In other words, Test-2 contains many distracting satellite images, and it considers the real deployment scenario compared to Test-1. Visual illustrations of the differences between Test-1 and Test-2 are provided in the supplementary material.
Tab.~\ref{tab:dataset} presents the query ground image numbers of the Training, Validation, Test-1, and Test-2 sets. 

\begin{table}[t!]
\setlength{\abovecaptionskip}{0pt}
    \setlength{\belowcaptionskip}{0pt}
\setlength{\tabcolsep}{8pt}
\centering
\caption{\scriptsize Query image numbers in the Training, Validation and Tests sets.}
\scriptsize{
\begin{tabular}{c|c|c|cc}
\toprule
           & Training & Validation & Test-1 & Test-2 \\ \midrule
Distractor & \xmark      &\xmark&\xmark  &\cmark  \\
Query Num  &  23,905     & 2,362    &  2,473     &  2,473    \\ \bottomrule
\end{tabular}}
\label{tab:dataset}
\vspace{-1em}
\end{table}

\section{Experiments} 

\noindent\textbf{Evaluation metrics.}
Following the previous cross-view localization work~\cite{Liu_2019_CVPR}, we use the distance and recall at top $k$ ($r@k$) for the performance evaluation. 
Specifically, when one of the retrieved top $k$ reference images is within $10$ meters to the query ground location, it is regarded as a successful localization. 
The percentage of successfully localized query images is recorded as recall at top $k$. 
we set $k$ to $1$, $5$, $10$ and $100$, respectively. 


\smallskip
\noindent\textbf{Implementation details. }
The input satellite image size is $512 \times 512$, center cropped from the collected images. The coverage of them is approximately $102\text{m} \times 102\text{m}$. 
The ground image resolution is $256 \times 1024$. 
The sizes of our global descriptor for query videos and satellite images are both $4096$, which is a typical descriptor dimension in image retrieval.  
We follow prior arts \cite{Hu_2018_CVPR,Liu_2019_CVPR,Regmi_2019_ICCV,Cai_2019_ICCV,shi2019spatial,shi2020optimal,shi2020looking,zhu2021revisiting,toker2021coming,zhu2021vigor} to adopt an exhaustive mini-batch strategy~\cite{vo2016localizing} with a batch size of $B=8$ to prepare the training triplets.
The Adam optimizer~\cite{kingma2014adam} with a learning rate of $10^{-4}$ is employed, and our network is trained end-to-end with five epochs. 
Our source code with every detail will be released, and the satellite images will be available for research purposes only and upon request. 

\begin{table*}[t!]
\setlength{\abovecaptionskip}{0pt}
\setlength{\belowcaptionskip}{0pt}
\setlength{\tabcolsep}{2pt}
\centering
\scriptsize
\caption{\scriptsize  Performance comparison on different designs for view projection and sequence fusion (sequence = 4)}
\begin{tabular}{c|c|c|HHHHcccc|cccc}
\toprule
                                                                                    &        & \multirow{2}{*}{\begin{tabular}[c]{@{}c@{}}Model\\ Size\end{tabular}}  & \multicolumn{4}{H}{Validation}                                           & \multicolumn{4}{c|}{Test-1}                                         & \multicolumn{4}{c}{Test-2}                                         \\
                                                                                    &        &            & r@1            & r@5            & r@10           & r@100          & r@1            & r@5            & r@10           & r@100          & r@1            & r@5            & r@10           & r@100          \\ \midrule
\multirow{2}{*}{\begin{tabular}[c]{@{}c@{}}View\\ Projection\end{tabular}} & Ours w/o GVP (Unet)   & 66.4M                       & 22.37          & 40.25          & 52.16          & 79.49          & 0.08           & 0.61           & 1.70           & 26.24          & 0.00           & 0.00           & 0.00           & 1.09           \\
& Ours w/o GVP          & 66.2M                       & 24.32          & 42.50          & 55.30          & 90.47          & 1.66           & 4.33           & 7.97           & 36.35          & 0.04           & 0.16           & 0.20           & 5.22           \\ \midrule
\multirow{3}{*}{\begin{tabular}[c]{@{}c@{}}Direct\\ Fusion\end{tabular}}                                                      & Conv2D & 66.0M      & 36.99         & 61.61         & 76.78         & 98.09         & 1.25           & 5.90           & 10.80          & 65.91          & 8.90           & 18.44          & 26.61          & 76.51          \\
                                                                                    & Conv3D & 66.0M      & 37.37          & 59.45          & 72.92          & 96.53          & 15.08          & 41.57          & 53.17          & 93.09          & 7.00           & 20.38          & 30.33          & 75.90          \\
                                                                                    & LSTM   & 66.2M      & 28.22          & 53.39          & 69.66          & 96.65          & 12.53          & 32.11          & 50.42          & 96.93          & 5.78           & 15.89          & 23.01          & 70.60          \\ \midrule
\multirow{4}{*}{\begin{tabular}[c]{@{}c@{}}Attention \\based\\  Fusion\end{tabular}} & Conv2D & 66.0M      & \textbf{45.89} & \textbf{69.58} & \textbf{82.88} & 96.61          & 18.80          & 47.03          & 61.75          & 96.64          & 11.69          & 25.03          & 36.55          & 81.52          \\
                                                                                    & Conv3D & 66.0M      & 42.80          & 67.03          & 79.36          & 95.13          & 19.65          & 43.27          & 58.39          & 97.41          & 11.36          & 24.02          & 34.45          & 83.58          \\
                                                                                    & LSTM   & 66.1M      & 40.21          & 63.26          & 75.68          & 93.47          & 15.93          & 47.88          & \textbf{66.03} & 97.61          & 9.70           & 24.30          & 35.26          & \textbf{85.08} \\
                                                                                    & \textbf{Ours}   & 66.2M      & { 42.58}    & { 66.65}    & { 79.66}    & \textbf{98.52} & \textbf{21.80} & \textbf{47.92} & { 64.94}    & \textbf{99.07} & \textbf{12.90} & \textbf{27.34} & \textbf{38.62} & { 85.00}    \\ \bottomrule
\end{tabular}
\label{tab:sequence_fusion}
\vspace{-1em}
\end{table*}

\subsection{Cross-view video-based localization}
Since there are no existing video-based cross-view localization algorithms, we conduct extensive experiments to dissect the effectiveness and necessity of each component in our framework. 

\subsubsection{Geometry-driven view projection.}
Although our GVP module is the basis for the following sequence fusion and similarity matching steps, we investigate whether it can be replaced or removed. 
We first replace it with an Unet and expect the domain correspondences can be learned implicitly during training, denoted as ``Ours w/o GVP (Unet)''. 
Next, we remove it from our pipeline and directly feed the original ground-view features to our sequence fusion module, denoted as ``Ours w/o GVP''. 
As indicated by the results in Tab.~\ref{tab:sequence_fusion}, the performance of the two baselines is significantly inferior to our whole pipeline, demonstrating the necessity of our geometry-driven view projection module. 

\begin{figure*}[ht!]
\setlength{\abovecaptionskip}{0pt}
    \setlength{\belowcaptionskip}{0pt}
    \centering
    \subfigure[\scriptsize Sequential ground-view images (sampled)]{
    \centering
    \begin{minipage}{0.48\linewidth}
    \includegraphics[width=0.49\linewidth, height=0.2037\linewidth]{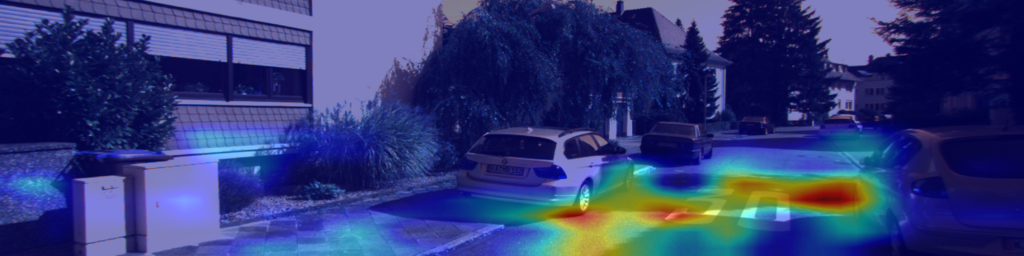}
    \includegraphics[width=0.49\linewidth, height=0.2037\linewidth]{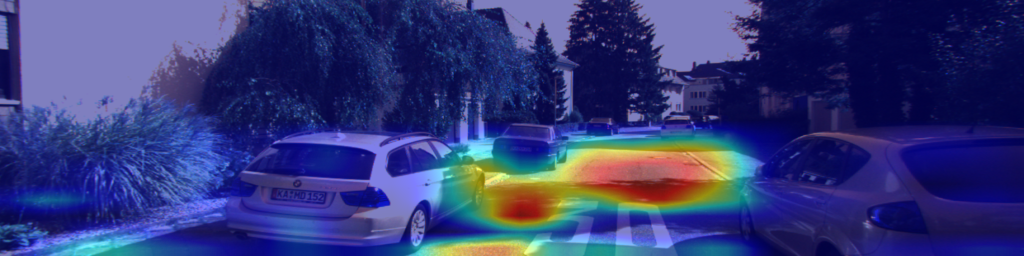}\\
    \includegraphics[width=0.49\linewidth, height=0.2037\linewidth]{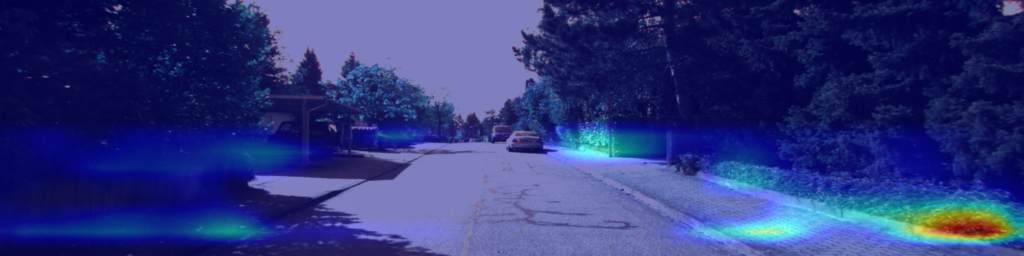}
    \includegraphics[width=0.49\linewidth, height=0.2037\linewidth]{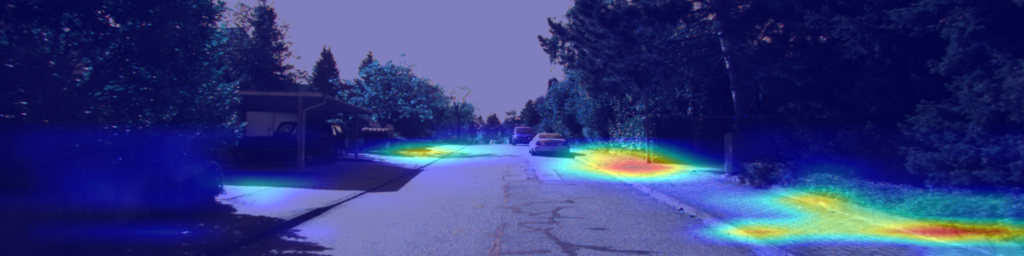}\\
    \includegraphics[width=0.49\linewidth, height=0.2037\linewidth]{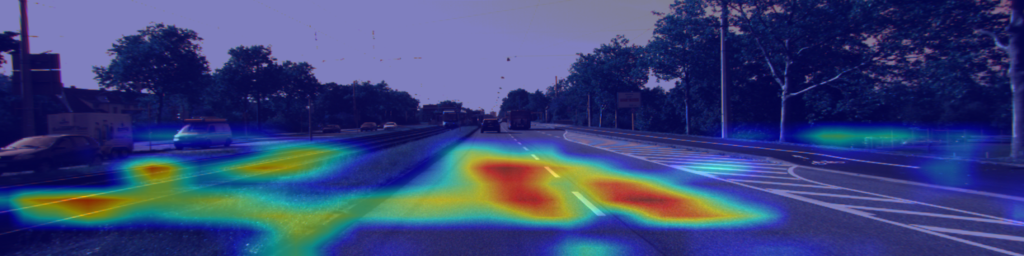}
    \includegraphics[width=0.49\linewidth, height=0.2037\linewidth]{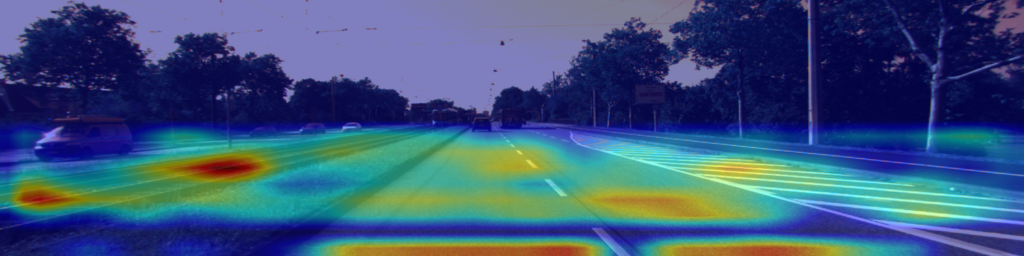} \\
    \includegraphics[width=0.49\linewidth, height=0.2\linewidth]{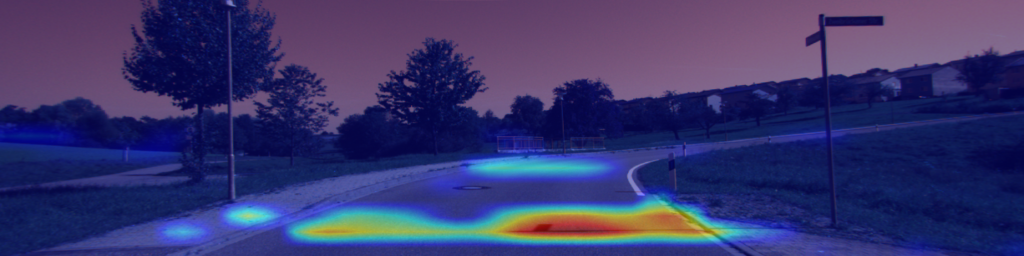}
    \includegraphics[width=0.49\linewidth, height=0.2\linewidth]{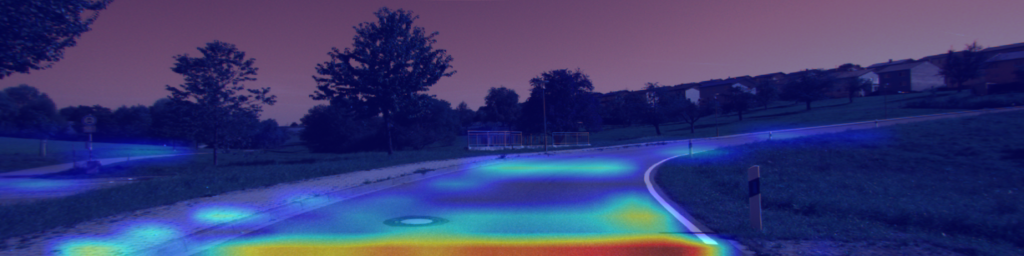}
    \end{minipage}
    \label{subfig:grd}
    }
    \subfigure[\scriptsize Satellite image]{
    \centering
    \begin{minipage}{0.096\linewidth}
    \includegraphics[width=\linewidth]{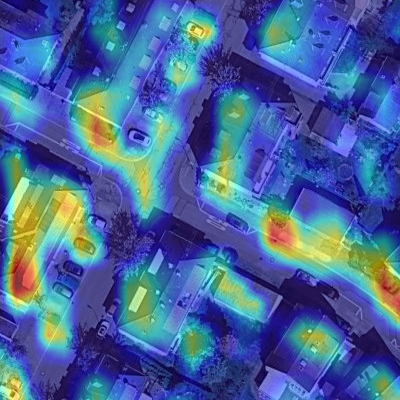} \\
    \includegraphics[width=\linewidth]{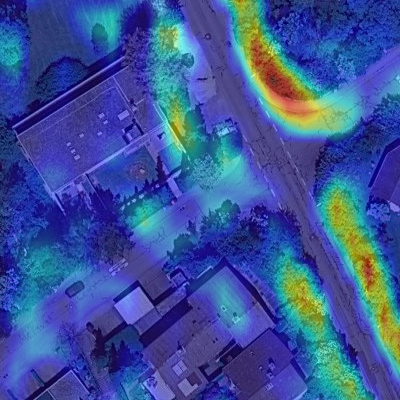} \\
    \includegraphics[width=\linewidth]{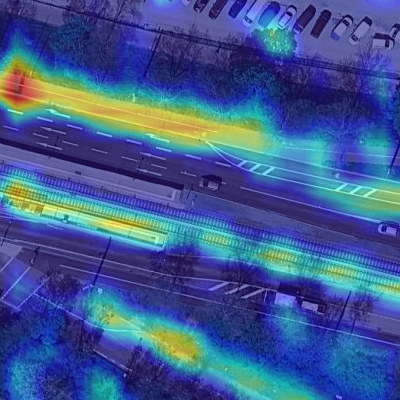} \\
    \includegraphics[width=\linewidth]{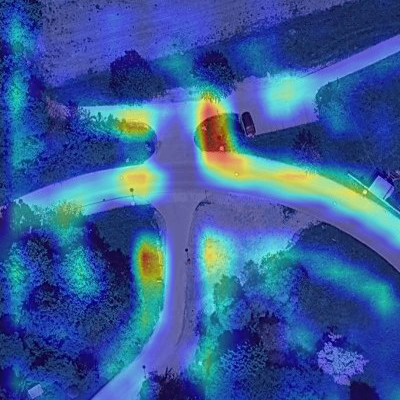} 
    \end{minipage}
    \label{subfig:sat}
    }
    \subfigure[\scriptsize Query feature]{
    \centering
    \begin{minipage}{0.096\linewidth}
    \includegraphics[width=\linewidth]{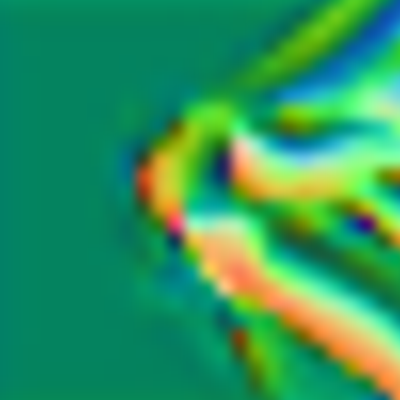}\\
    \includegraphics[width=\linewidth]{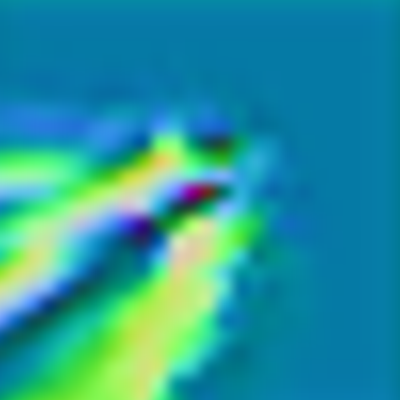}\\
    \includegraphics[width=\linewidth]{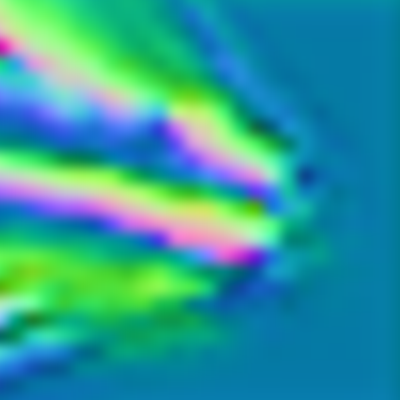} \\
    \includegraphics[width=\linewidth]{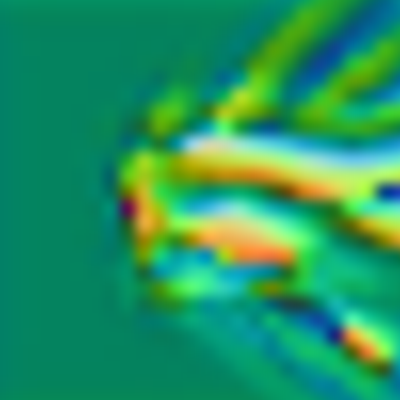}
    \end{minipage}
    \label{subfig:grd_feat}
    }
    \subfigure[\scriptsize Satellite feature]{
    \centering
    \begin{minipage}{0.096\linewidth}
    \includegraphics[width=\linewidth]{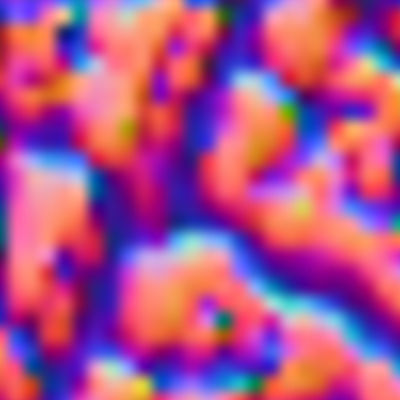}\\
    \includegraphics[width=\linewidth]{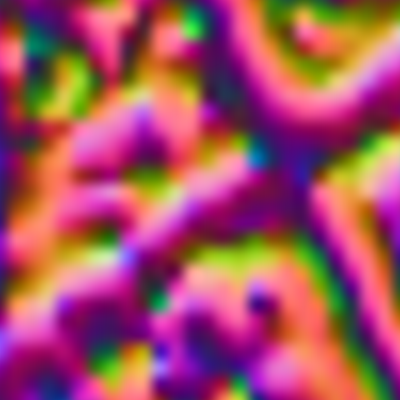}\\
    \includegraphics[width=\linewidth]{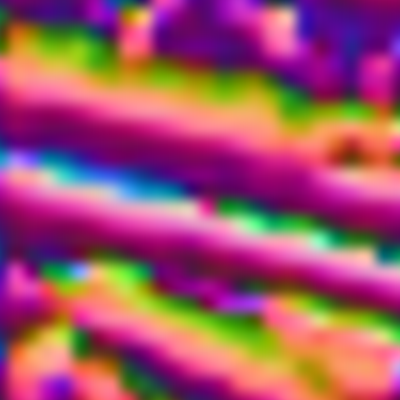} \\ 
    \includegraphics[width=\linewidth]{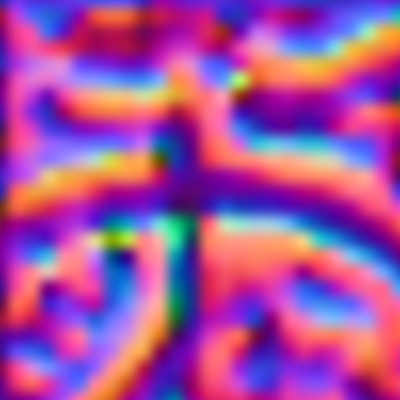}
    \end{minipage}
    \label{subfig:sat_feat}
    }
    \subfigure[\scriptsize Confidence map]{
    \centering
    \begin{minipage}{0.096\linewidth}
    \includegraphics[width=\linewidth]{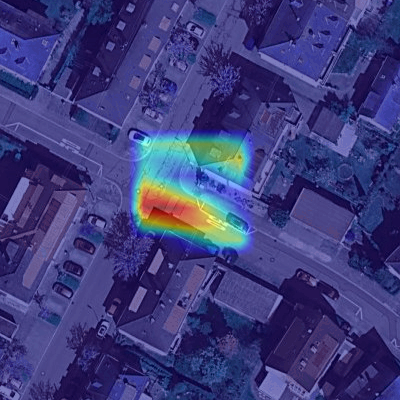}\\
    \includegraphics[width=\linewidth]{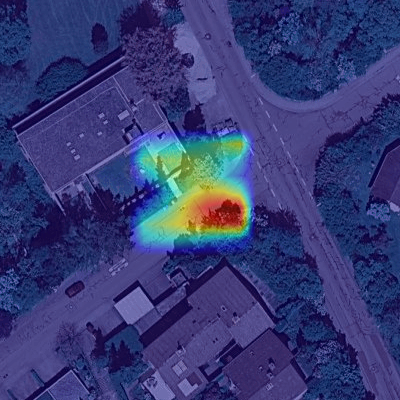}\\
    \includegraphics[width=\linewidth]{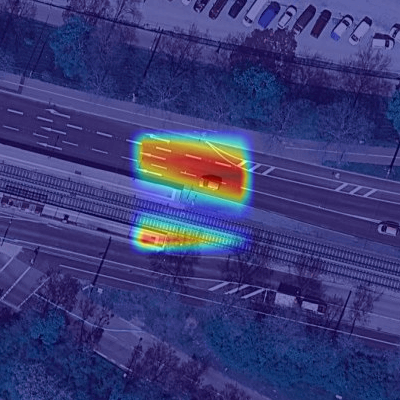}\\
    \includegraphics[width=\linewidth]{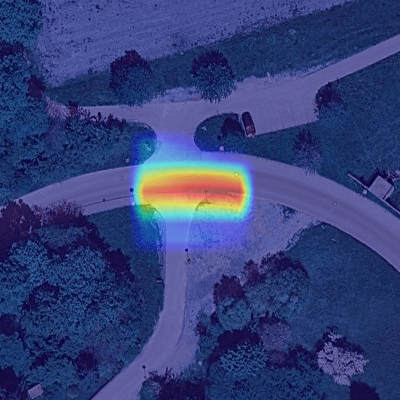}
    \end{minipage}
    \label{subfig:conf}
    }
    \caption{\scriptsize  Visualizations of intermediate results of our method. For ground images, the learned activations focus on salient features of the ground and road edges. Interestingly, it automatically ignores the dynamic objects (first row in (a)). The learned activations of satellite images concentrate on the scene objects along the main road (likely visible by a moving vehicle). The fused query video features (c) capture scene objects similar to those of their satellite counterparts (d), and the learned confidence maps attention on the region of road. 
    }
    \label{fig:visualization}
    \vspace{-1em}
\end{figure*}

\smallskip
\noindent 
\textbf{Learned viewpoint-invariant semantic features.}
To fully understand the capability of our view projection module, we visualize the learned viewpoint-invariant semantic features of our network by using the techniques of Grad-Cam~\cite{Selvaraju_2017_ICCV}. 
As seen in Fig.~\ref{fig:visualization}, salient features on roads and roads edges are successfully recognized in ground-view images~(Fig.~\ref{subfig:grd}). 
The detected salient features in satellite images also concentrate on roads and scene objects along roads edges~(Fig.~\ref{subfig:sat}). 
By using our view projection module and the photo-consistency constrained sequence fusion mechanism, the learned global representations of the ground video~(Fig.~\ref{subfig:grd_feat}) capture similar scene patterns to those of their matching satellite counterparts~(Fig.~\ref{subfig:sat_feat}).

\subsubsection{Photo-consistency constrained sequence fusion.}
Our goal is to synthesize an overhead-view feature map from a query ground video. 
To this end, our PCSF module measures the photo consistency for each overhead view pixel across different ground-view images and fuses them with an attention-based (transformer) architecture. 
Apart from this design, LSTM (RNNs) and 3D CNNs are also known for their power to handle sequential signals. 
Hence, we compare with these architectures. 
For completeness, we also experiment with 2D CNNs.

\smallskip
\noindent
\textbf{Direct fusion.}
We first replace our PCSF module with Conv2D, Conv3D, and LSTM based networks, respectively. 
The Conv2D-based fusion network takes the projected sequential ground-view features $\mathbf{F}_{t_i}^p$ separately and computes the average of the outputs of different time steps. 
The Conv3D-based fusion network uses its third dimension to operate on the temporal dimension. The LSTM-based network includes two bidirectional LSTM layers to enhance the sequential relationship encoding. 
The outputs of the Conv3D-based and the LSTM-based networks are both directly fused features for the query video. 
The results are presented in the middle part of Tab.~\ref{tab:sequence_fusion}. 
It can be seen that the performance is significantly inferior to ours. 

\smallskip
\noindent
\textbf{Attention-based fusion. }
Based on the above observations, we infer that it may be difficult for a network to fuse a sequence of features implicitly. 
Hence, we employ the Conv2D, Conv3D, and LSTM based network to regress the attention weights for the projected features at different time steps, 
denoted as $\mathbf{N}_{t_i} \in \mathbb{R} ^ {S\times S}$. 
Then, the global query descriptor is obtained by a dot product between the attention weight $\mathbf{N}_{t_i}$ and the features $\mathbf{F}_{t_i}^p$. 
The results are presented in the bottom part of Tab.~\ref{tab:sequence_fusion}. 
It can be seen that the attention-based fusion methods all outperform the direct fusion methods, indicating that the attention-based decomposition helps to achieve better performance. 
Among the attention-based fusion ablations, our method achieves the best overall performance.
This should be attributed to the explicit photo consistency computation across different ground-views by our PCSF module.

\subsubsection{Different choices for network backbone.}

In this section, we conduct ablation study on different network backbones, including Vision transformer (ViT)~\cite{dosovitskiy2020image}, Swin transformer~\cite{liu2021swin}, Renet50~\cite{he2016deep} and VGG16~\cite{Simonyan2014VeryDC}(ours). 
Transformers are known of their superior feature extraction ability than CNNs. However, they do not preserve the translational equivariance ability, which however is an essential element in estimating the relative displacement between query camera locations and their matching satellite image centers. Thus, transformers achieve slightly worse performance than CNNs, as indicated by Tab.~\ref{tab:backbone}. Compared to VGG16, Resnet50 does not make significant improvement. 
Hence, following most previous works~\cite{Hu_2018_CVPR,shi2019spatial,shi2020optimal,shi2020looking,zhu2021vigor}, we use VGG16 as our network backbone.

\begin{table}[t!]
\setlength{\abovecaptionskip}{0pt}
\setlength{\belowcaptionskip}{0pt}
\setlength{\tabcolsep}{6pt}
\centering
\scriptsize
\caption{\scriptsize Performance comparison with different backbones (sequence = 4)
}
\begin{tabular}{l|cccc|cccc}
\toprule
\multicolumn{1}{c|}{} & \multicolumn{4}{c|}{Test-1}                                         & \multicolumn{4}{c}{Test-2}                                         \\
\multicolumn{1}{c|}{} & r@1            & r@5            & r@10           & r@100          & r@1            & r@5            & r@10           & r@100          \\ \midrule
ViT~\cite{dosovitskiy2020image}        & 20.05          & 45.13          & 60.17          & 97.53          & {{12.86}}   & {{27.94}}   & \textbf{38.86} & {{81.64}}          \\
Swin~\cite{liu2021swin}                & 18.40          & 47.80          & 63.73          & \textbf{99.11} & 12.29          & 22.31          & 35.29          & 80.70          \\
Resnet~\cite{he2016deep}               & \textbf{22.68} & \textbf{55.16} & \textbf{67.69} & {97.90}        & {9.75}         & \textbf{28.31} & {38.45}        & {73.72} \\
VGG16~\cite{Simonyan2014VeryDC} (Ours) & {{21.80}}   & {{47.92}}   & {{64.94}}   & {{99.07}}  & \textbf{12.90} & {27.34}         & {{38.62}}     & \textbf{ 85.00}     \\ \bottomrule
\end{tabular}
\label{tab:backbone}
\vspace{-0.5em}
\end{table}

\begin{table}[t!]
\setlength{\abovecaptionskip}{0pt}
\setlength{\belowcaptionskip}{0pt}
\setlength{\tabcolsep}{7pt}
\centering
\scriptsize
\caption{\scriptsize Effectiveness of the scene-prior driven similarity matching (sequence = 4)}
\begin{tabular}{l|cccc|cccc}
\toprule
\multicolumn{1}{c|}{} & \multicolumn{4}{c|}{Test-1}                                         & \multicolumn{4}{c}{Test-2}                                         \\
\multicolumn{1}{c|}{} & r@1            & r@5            & r@10           & r@100          & r@1            & r@5            & r@10           & r@100          \\ \midrule
Ours w/o SSM         & 6.35           & 25.76          & 41.97          & 97.61          & 3.48           & 9.42           & 14.03          & 63.04          \\
Ours w/o U           & { 13.26}    & { 36.76}    & { 55.72}    & { 97.05}    & { 10.47}    & \textbf{27.42} & \textbf{39.51} & \textbf{88.92} \\
Ours                 & \textbf{21.80} & \textbf{47.92} & \textbf{64.94} & \textbf{99.07} & \textbf{12.90} & { 27.34}    & { 38.62}    & { 85.00}     \\ \bottomrule
\end{tabular}
\label{tab:ablation}
\end{table}

\subsubsection{Scene-prior driven similarity matching.}

Next, we study whether the NCC-based similarity matching can be removed. 
In this experiment, the distance between the satellite features and the fused ground-view features is directly computed without estimating their potential alignments. Instead, they are assumed to be aligned at the satellite image center. 
The results are presented in the first row of Tab.~\ref{tab:ablation}. 
The performance drops significantly compared to our whole baseline, demonstrating that the network does not have the ability to tolerate the spatial shifts between query camera locations, and our explicit alignment strategy (NCC-based similarity matching) is effective. 

Furthermore, we investigate the effectiveness of the learned scene prior by the uncertainty map (Eq.~\ref{eq:ncc_uncertainty}).
To do so, we remove the term of uncertainty map $\mathbf{U}(\mathbf{F}^s)_{m, n}$ in Eq.~\eqref{eq:ncc_uncertainty}, denoted as ``Ours w/o U''. 
The results in the second row of Tab.~\ref{tab:ablation} indicates the learned uncertainty map boosts the localization performance. 
Fig.~\ref{subfig:conf} visualizes the generated confidence maps (inverse of uncertainty) by our method. 
It can be seen that the higher confidence regions mainly concentrate on roads, indicating that the confidence maps successfully encode the semantic information of satellite images and recognize the correct possible regions for a vehicle location. 

\begin{figure}[t]
\setlength{\abovecaptionskip}{0pt}
    \setlength{\belowcaptionskip}{0pt}
    \centering
    \begin{minipage}{0.33\linewidth}
\setlength{\abovecaptionskip}{0pt}
    \setlength{\belowcaptionskip}{0pt}
    \centering
    \scalebox{1.0}[1.0]{\includegraphics[trim={8mm 65mm 10mm 70mm}, clip, width=0.9\linewidth]{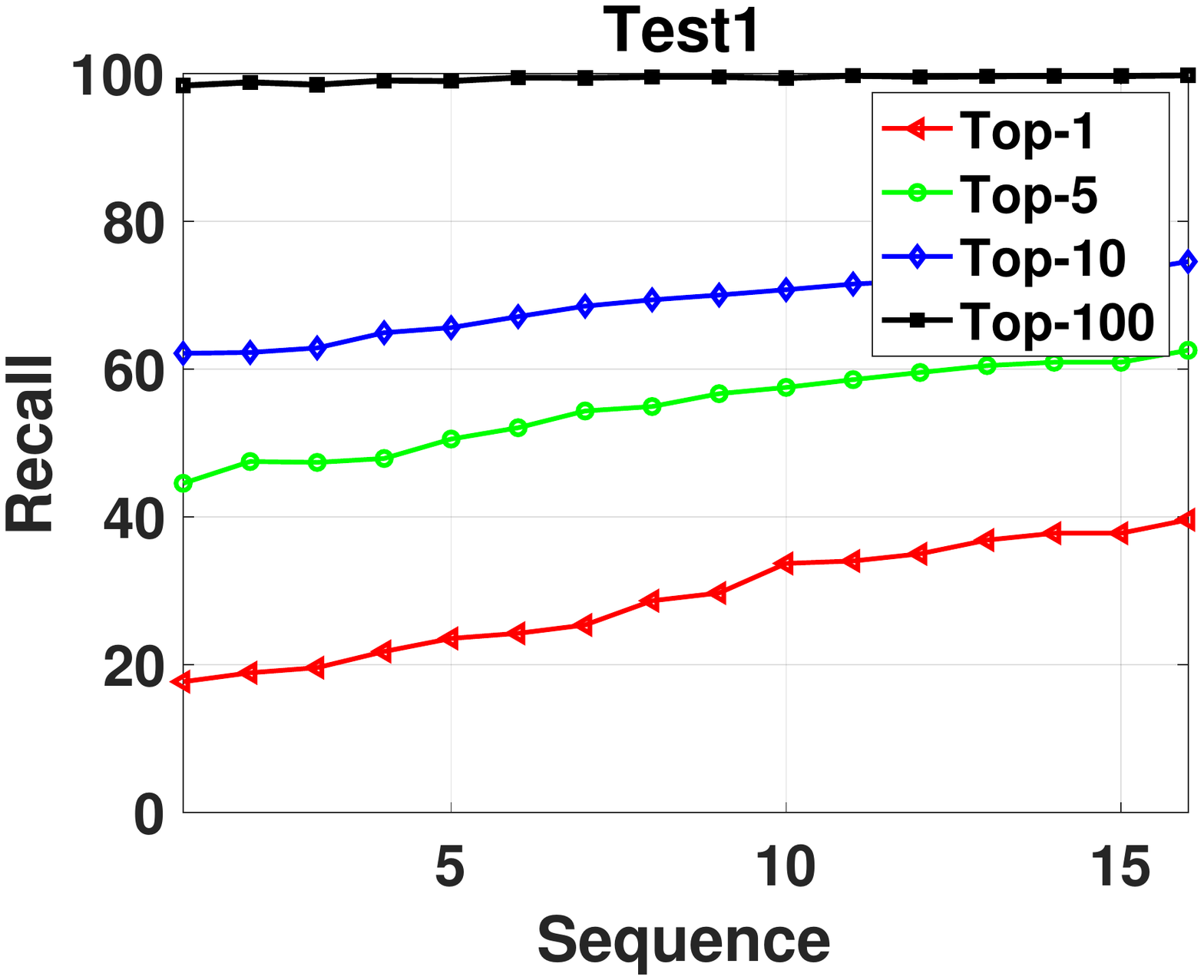}}\\
    \scalebox{1.0}[1.0]{\includegraphics[trim={8mm 65mm 10mm 70mm}, clip, width=0.9\linewidth]{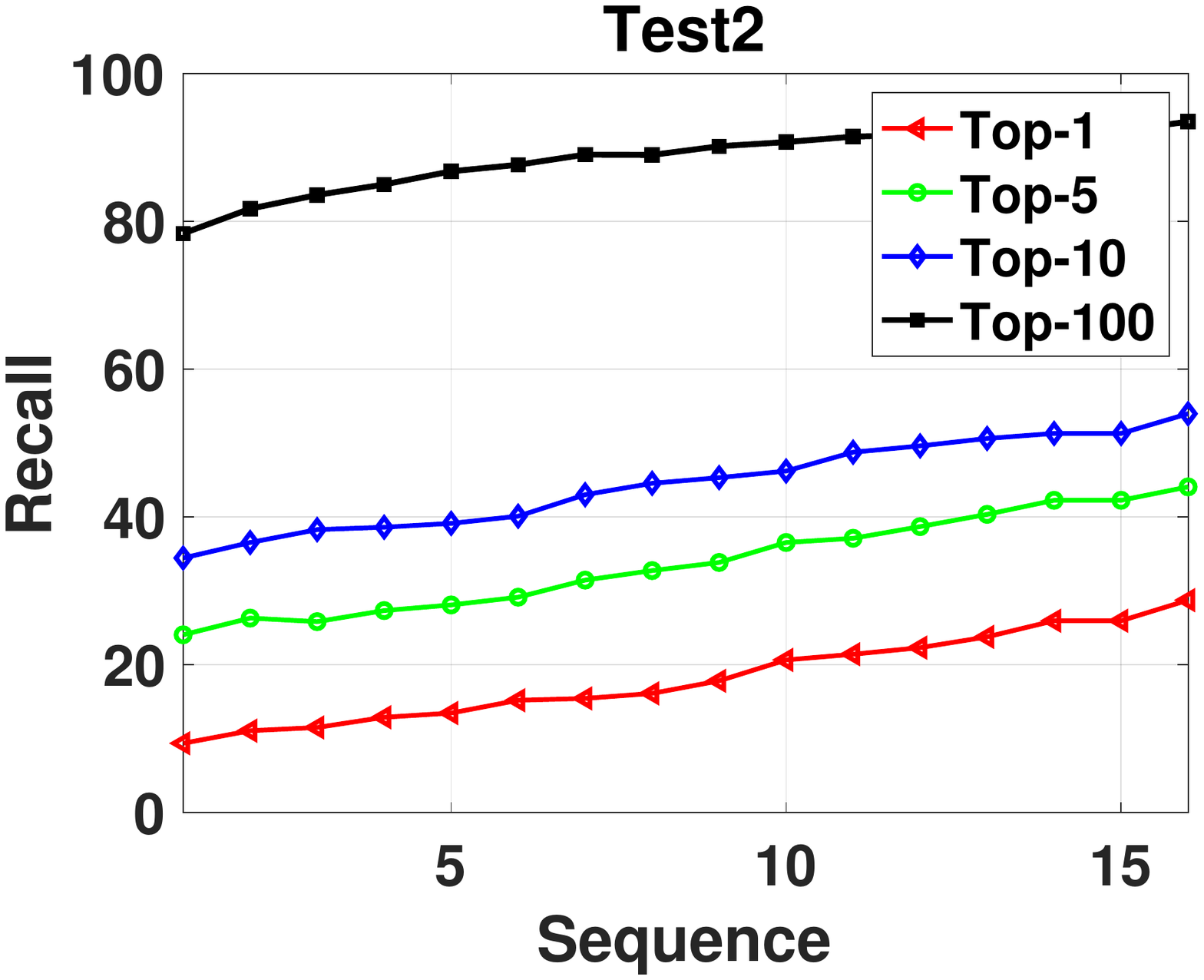}}
    \captionof{figure}{Recall rates with the increase of input sequence number.}
    \label{fig:sequence_length} 
\end{minipage}\hfill
\begin{minipage}{0.65 \linewidth}
\setlength{\abovecaptionskip}{0pt}
    \setlength{\belowcaptionskip}{0pt}
    \centering
    \subfigure[\scriptsize Query Video]{
    \vspace{2mm}
    \begin{minipage}{0.27\linewidth}
    \centerline{\scriptsize }
    \includegraphics[width=\linewidth, height=0.48\linewidth]{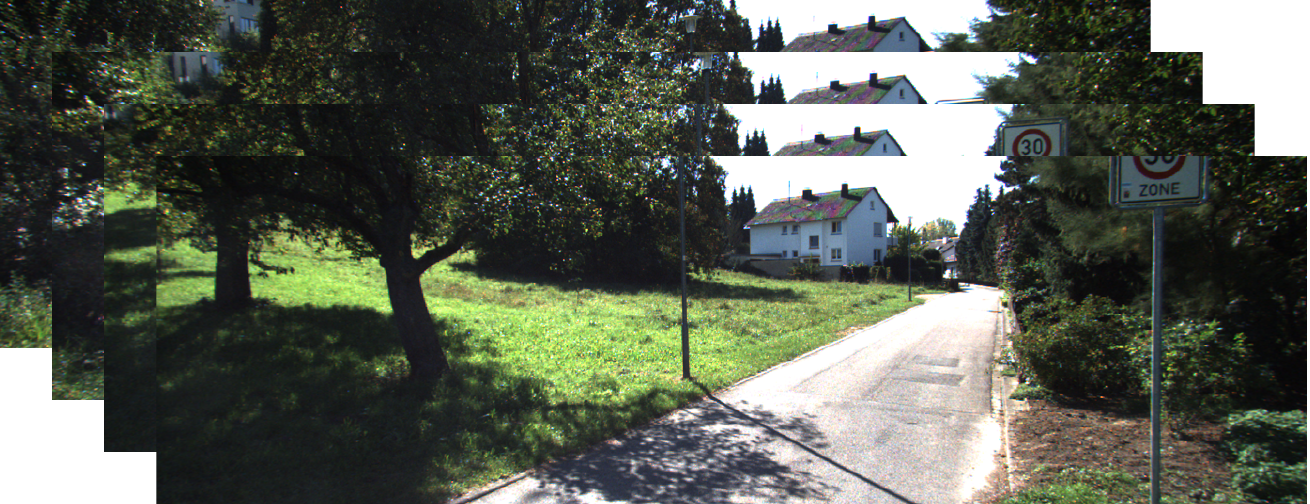}
    \centerline{\scriptsize }
    \includegraphics[width=\linewidth, height=0.48\linewidth]{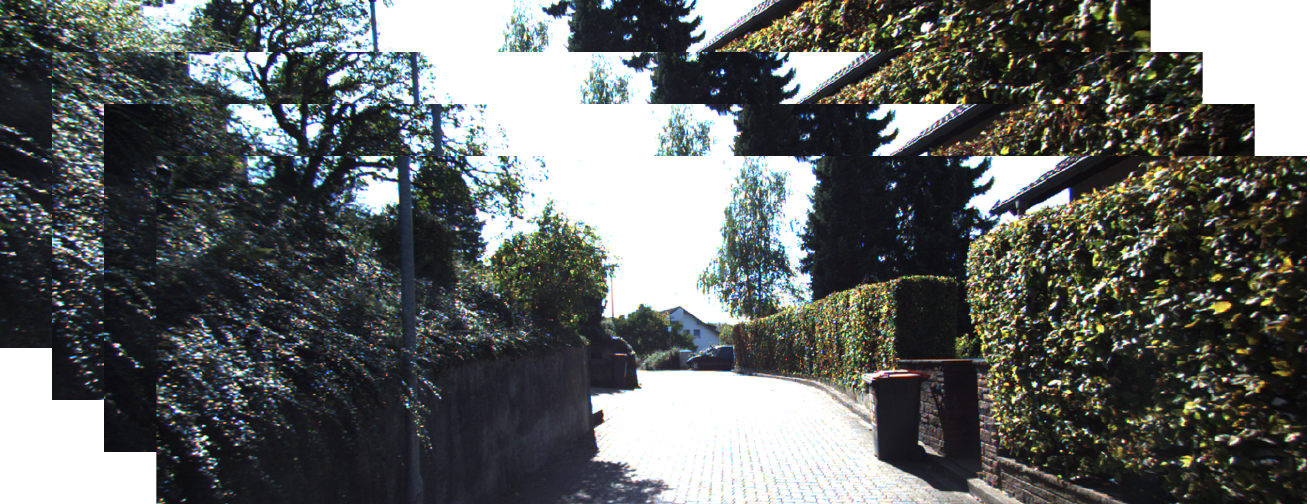}
    \centerline{\scriptsize }
    \includegraphics[width=\linewidth, height=0.48\linewidth]{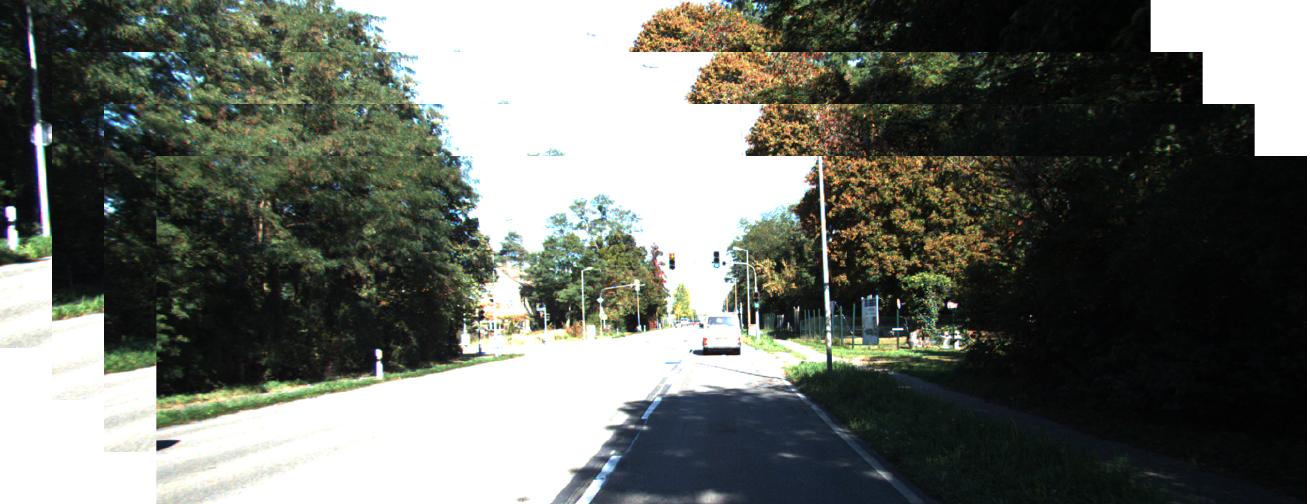}
    \centerline{\scriptsize }
    \includegraphics[width=\linewidth, height=0.48\linewidth]{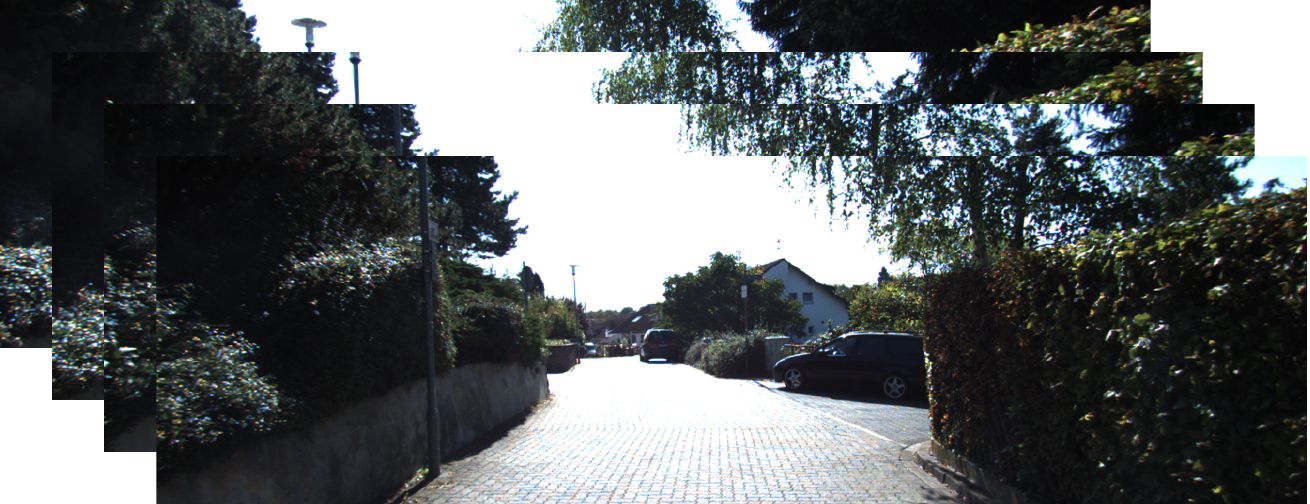}
    \end{minipage}
    }
    \subfigure[\scriptsize Top-1]{
    \begin{minipage}{0.13\linewidth}
    \centerline{\scriptsize 559 m}
    \includegraphics[width=\linewidth]{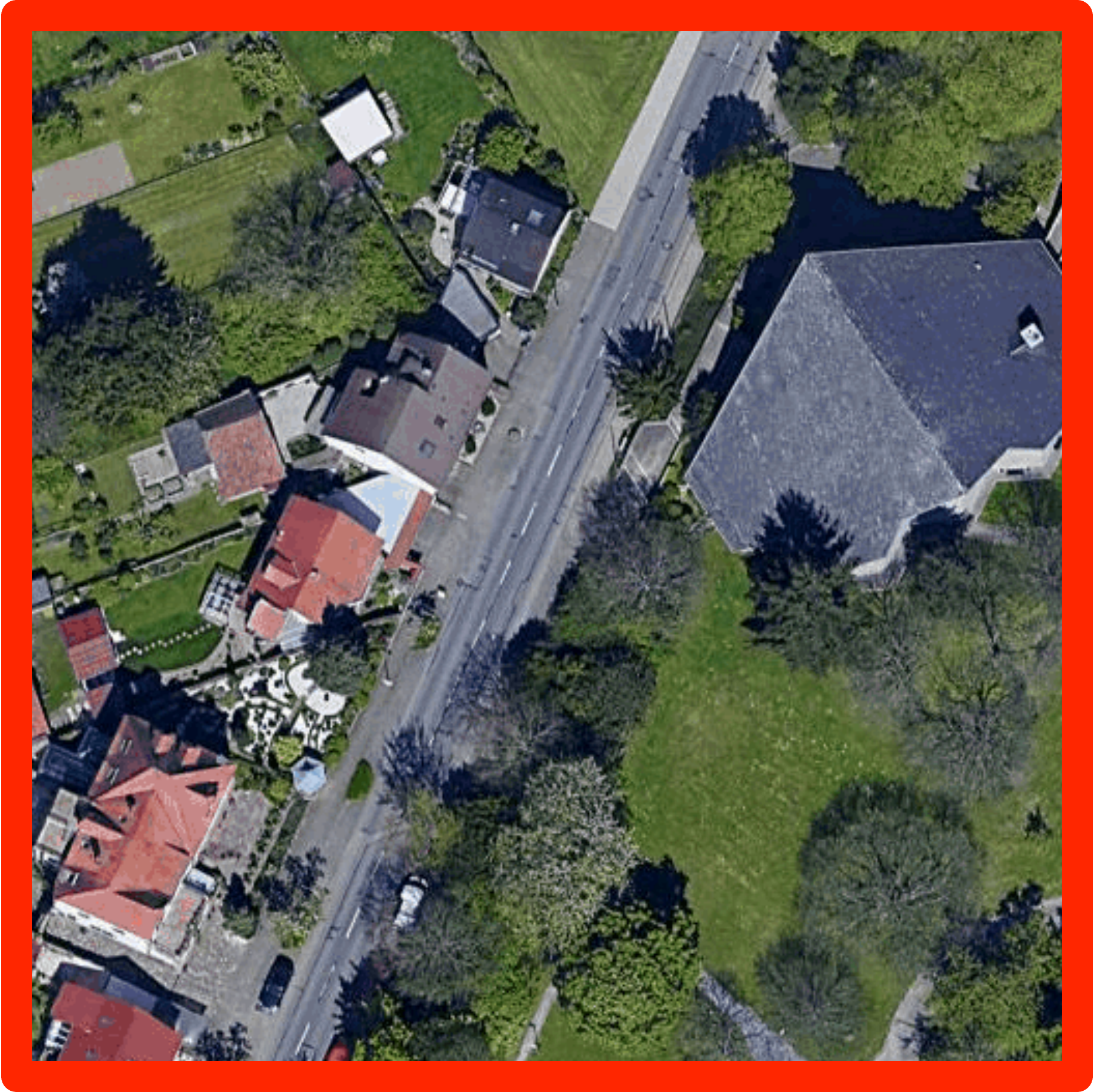}
    \centerline{\scriptsize 10811 m}
    \includegraphics[width=\linewidth]{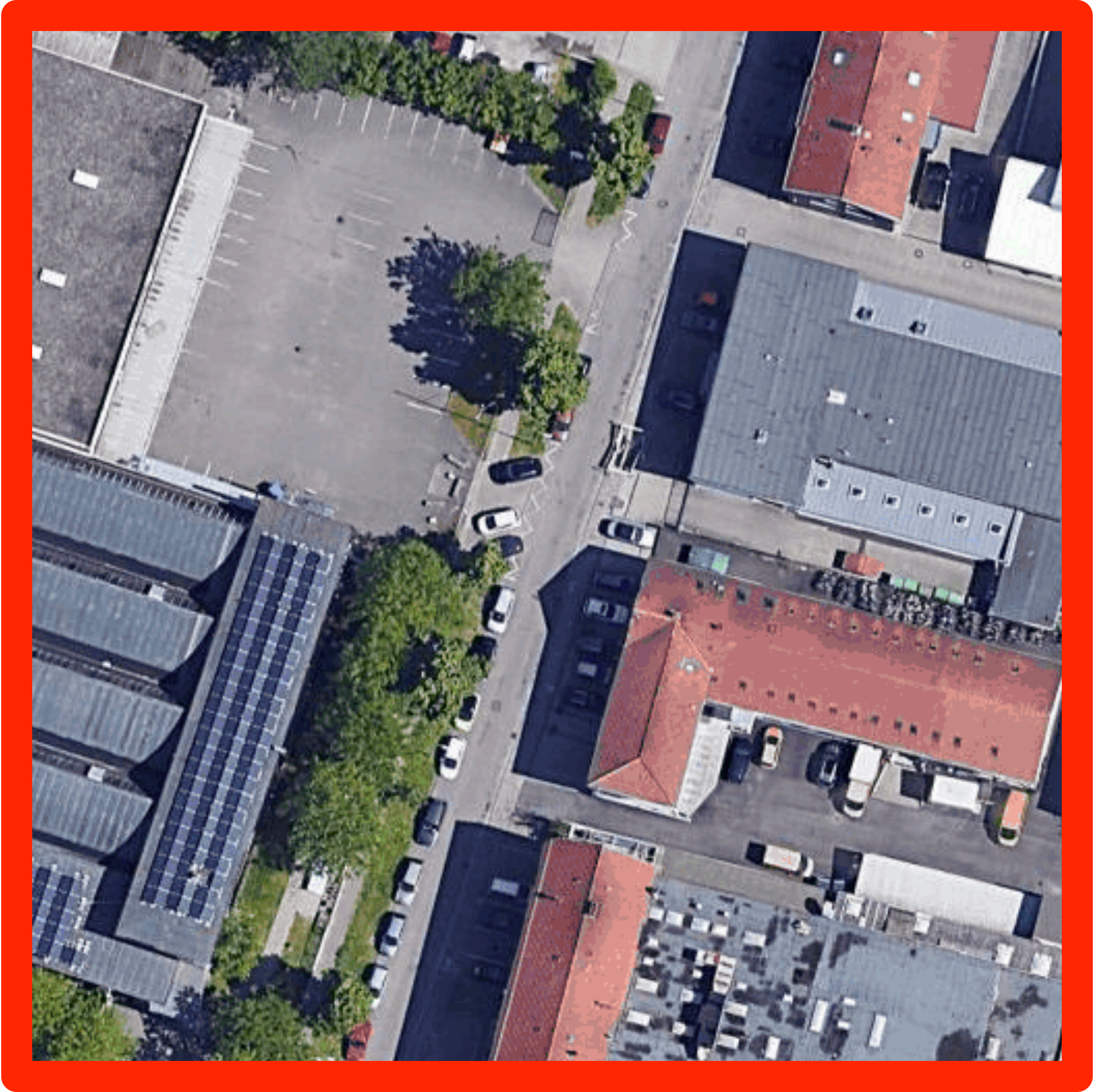}
    \centerline{\scriptsize 9479 m}
    \includegraphics[width=\linewidth]{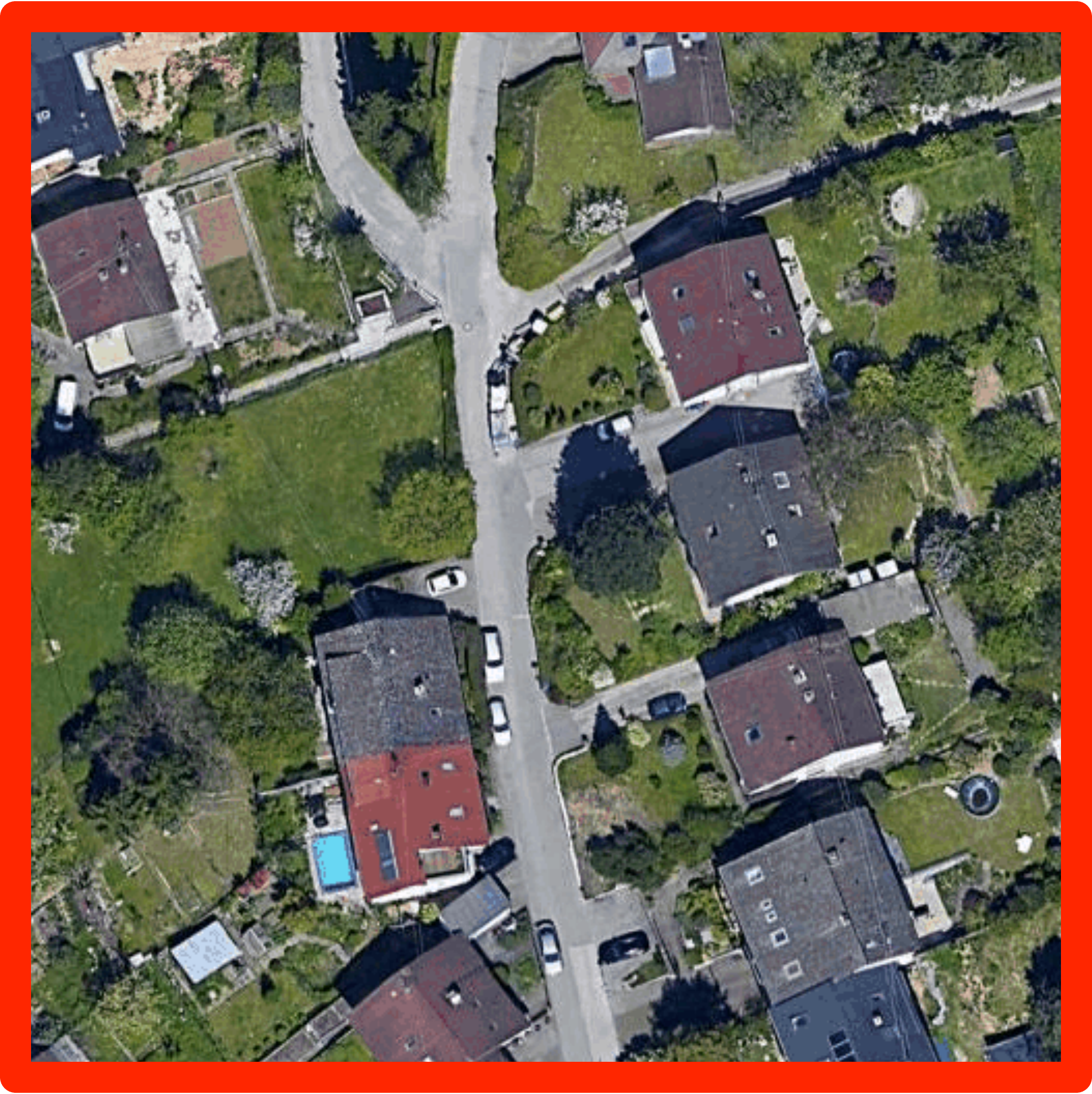}
    \centerline{\scriptsize 3 m}
    \includegraphics[width=\linewidth]{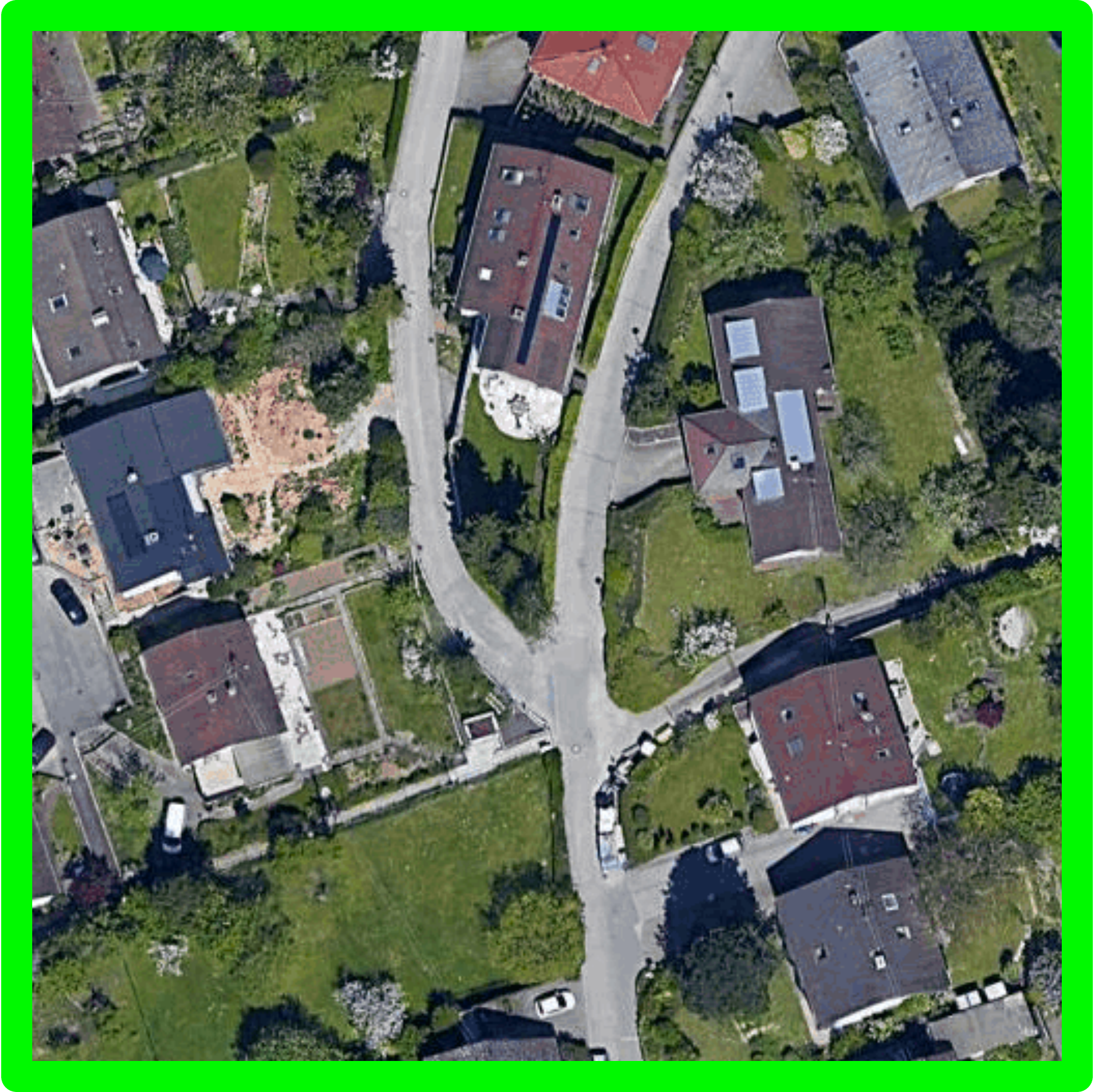}
    \end{minipage}
    }
    \subfigure[\scriptsize Top-2]{
    \begin{minipage}{0.13\linewidth}
    \centerline{\scriptsize 1 m}
    \includegraphics[width=\linewidth]{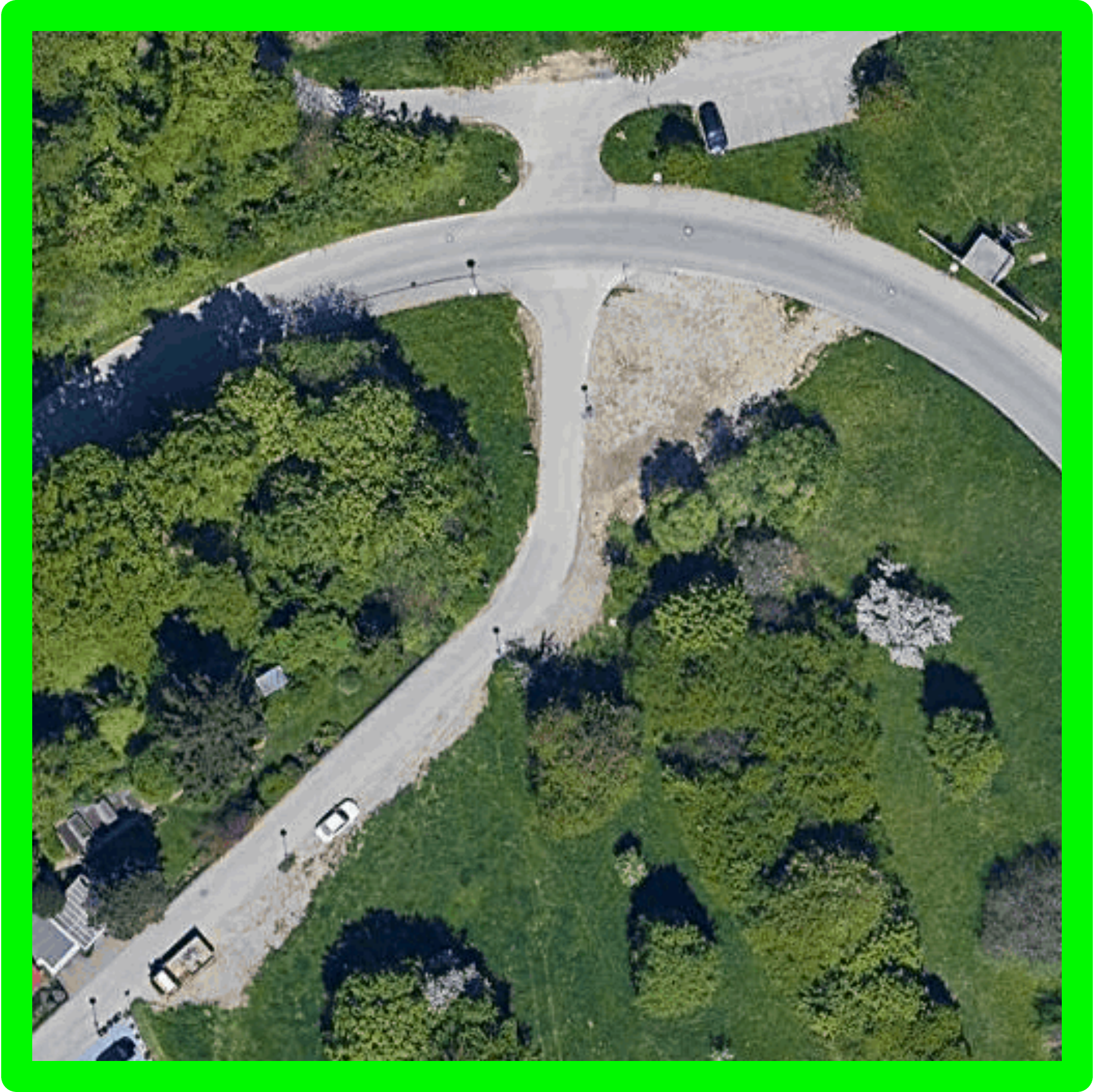}
    \centerline{\scriptsize 10802 m}
    \includegraphics[width=\linewidth]{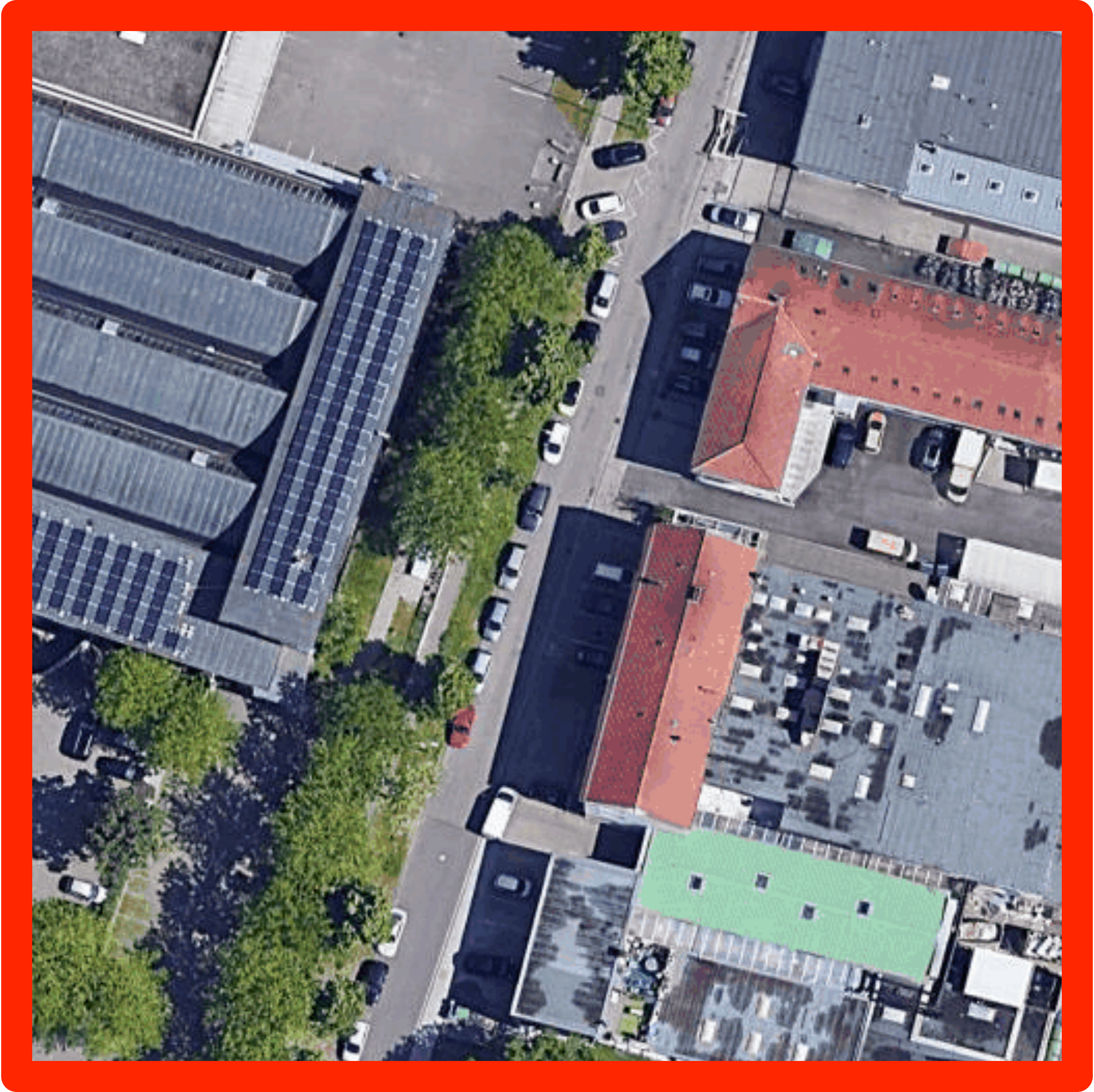}
    \centerline{\scriptsize 9363 m}
    \includegraphics[width=\linewidth]{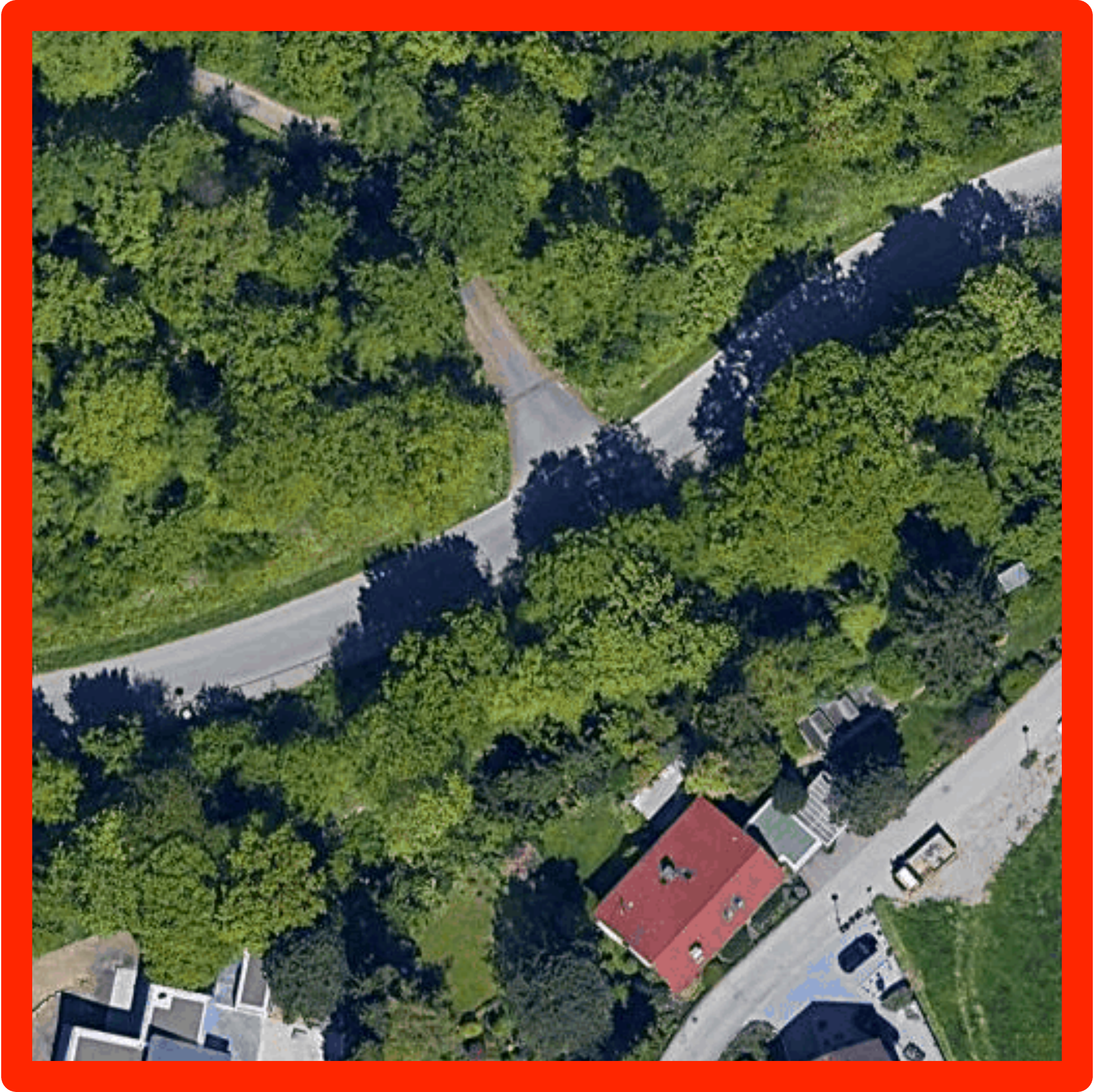}
    \centerline{\scriptsize 7 m}
    \includegraphics[width=\linewidth]{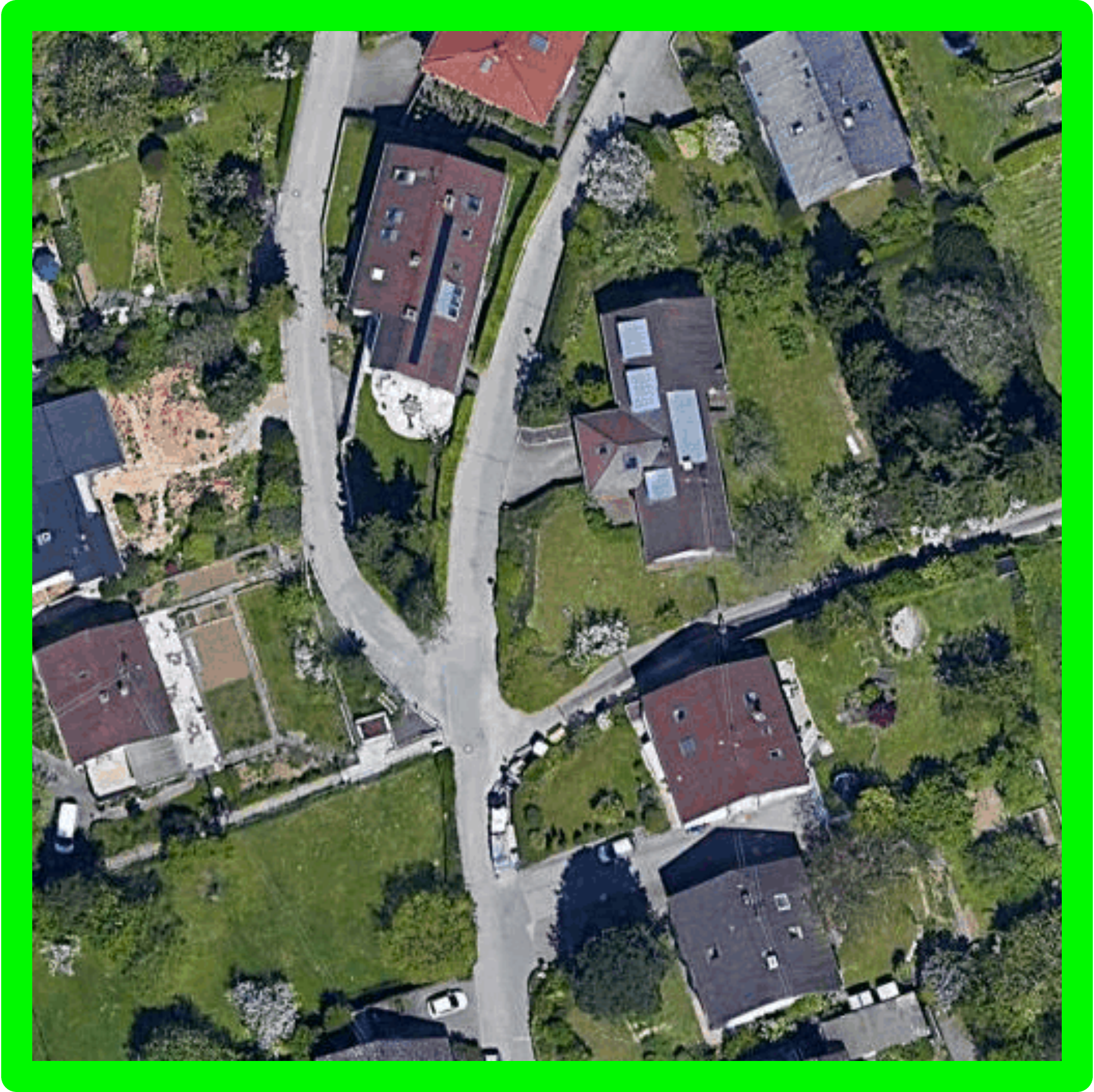}
    \end{minipage}
    }
    \subfigure[\scriptsize Top-3]{
    \begin{minipage}{0.13\linewidth}
    \centerline{\scriptsize 5990 m}
    \includegraphics[width=\linewidth]{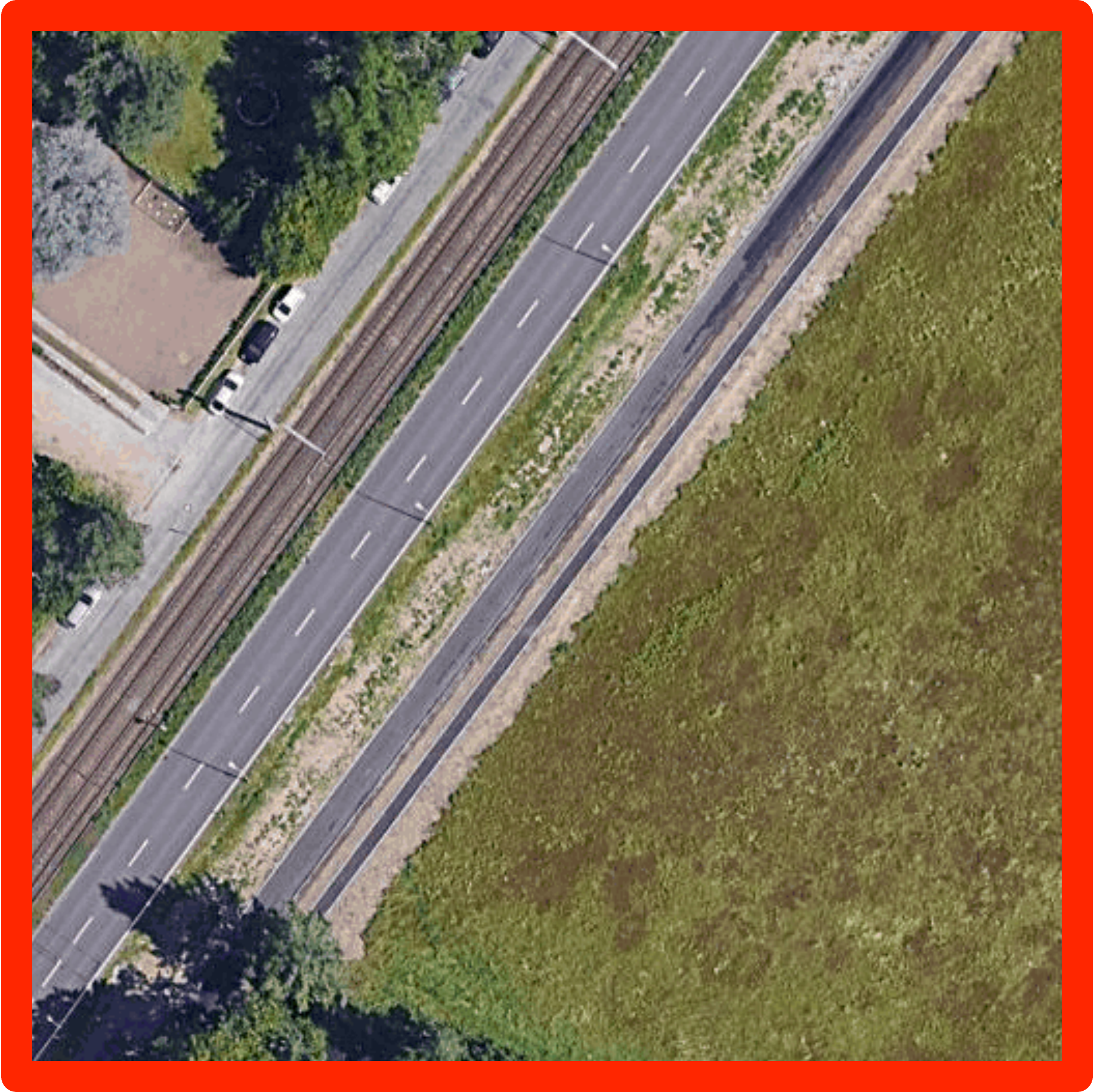}
    \centerline{\scriptsize 10808 m}
    \includegraphics[width=\linewidth]{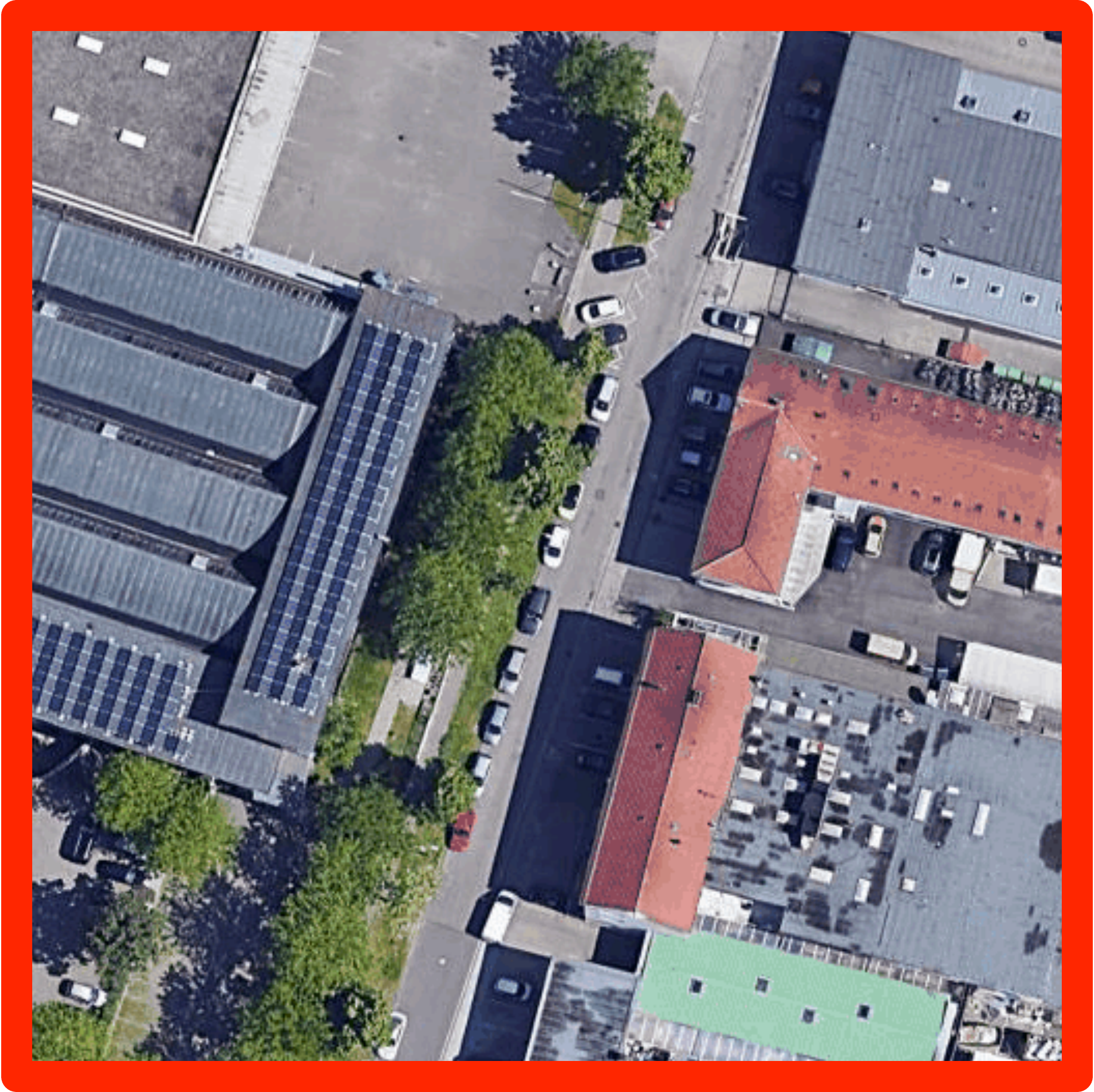}
    \centerline{\scriptsize 10901 m}
    \includegraphics[width=\linewidth]{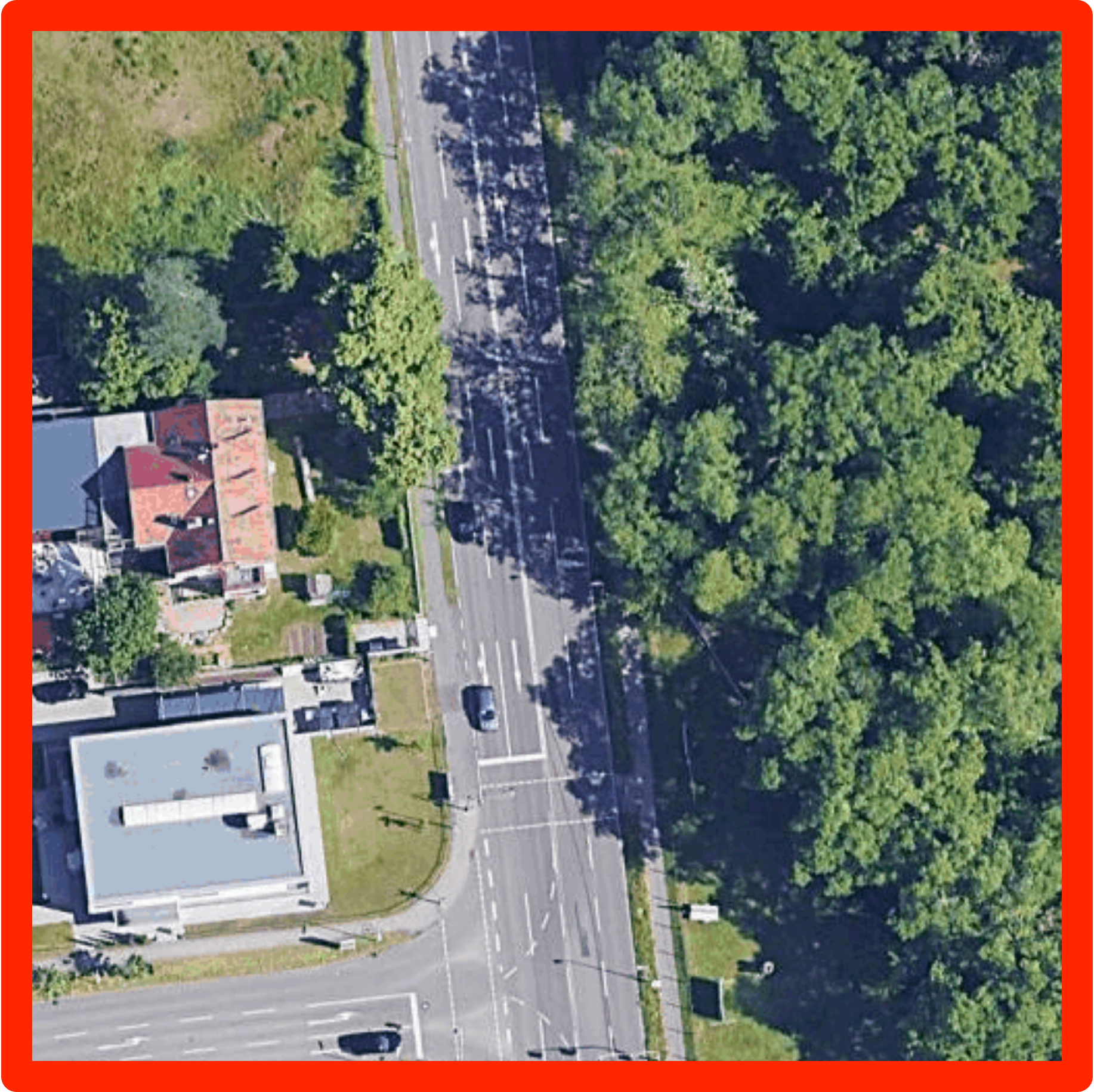}
    \centerline{\scriptsize 5990 m}
    \includegraphics[width=\linewidth]{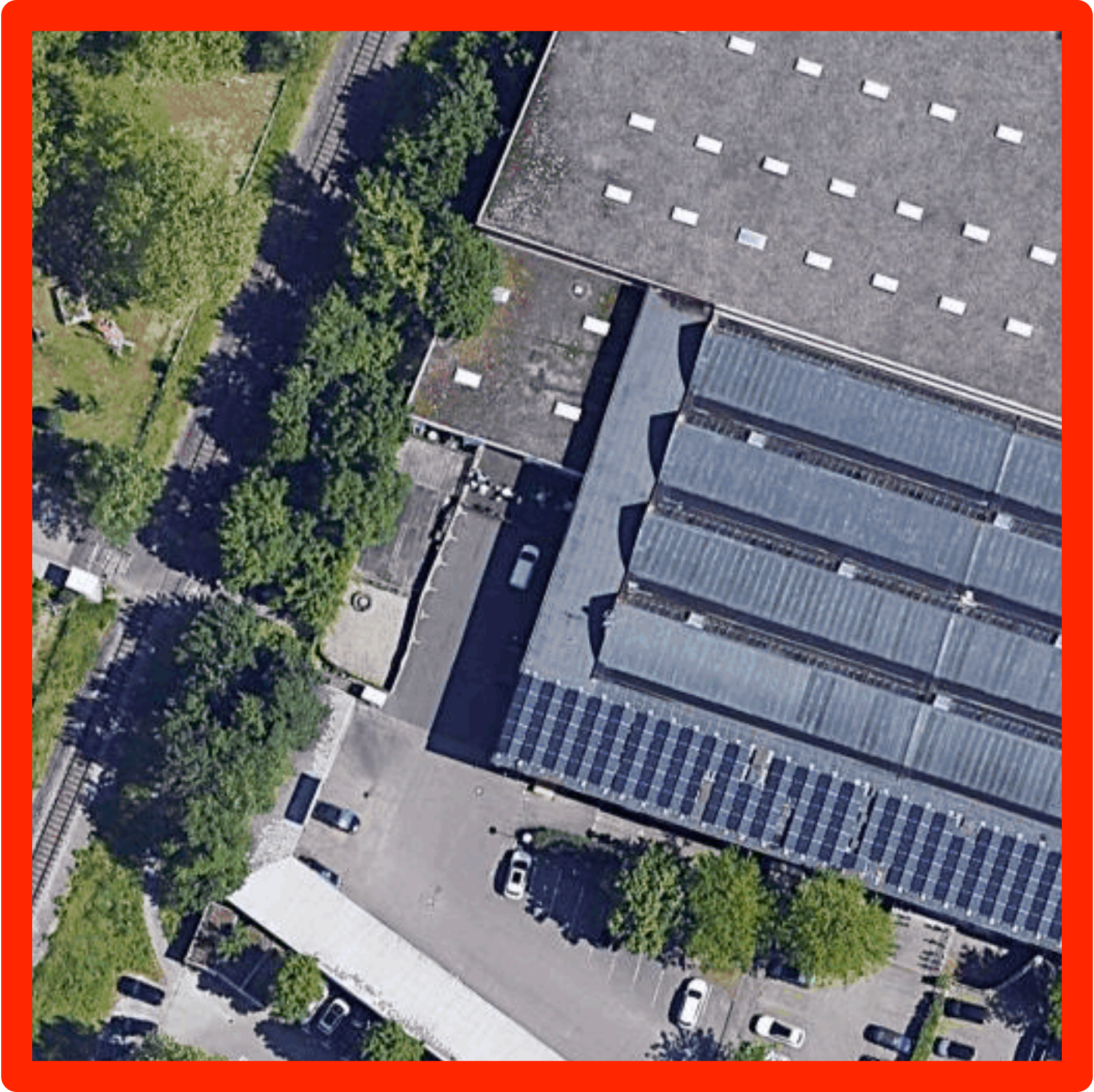}
    \end{minipage}
    }
    \subfigure[\scriptsize Top-4]{
    \begin{minipage}{0.13\linewidth}
    \centerline{\scriptsize 5995 m}
    \includegraphics[width=\linewidth]{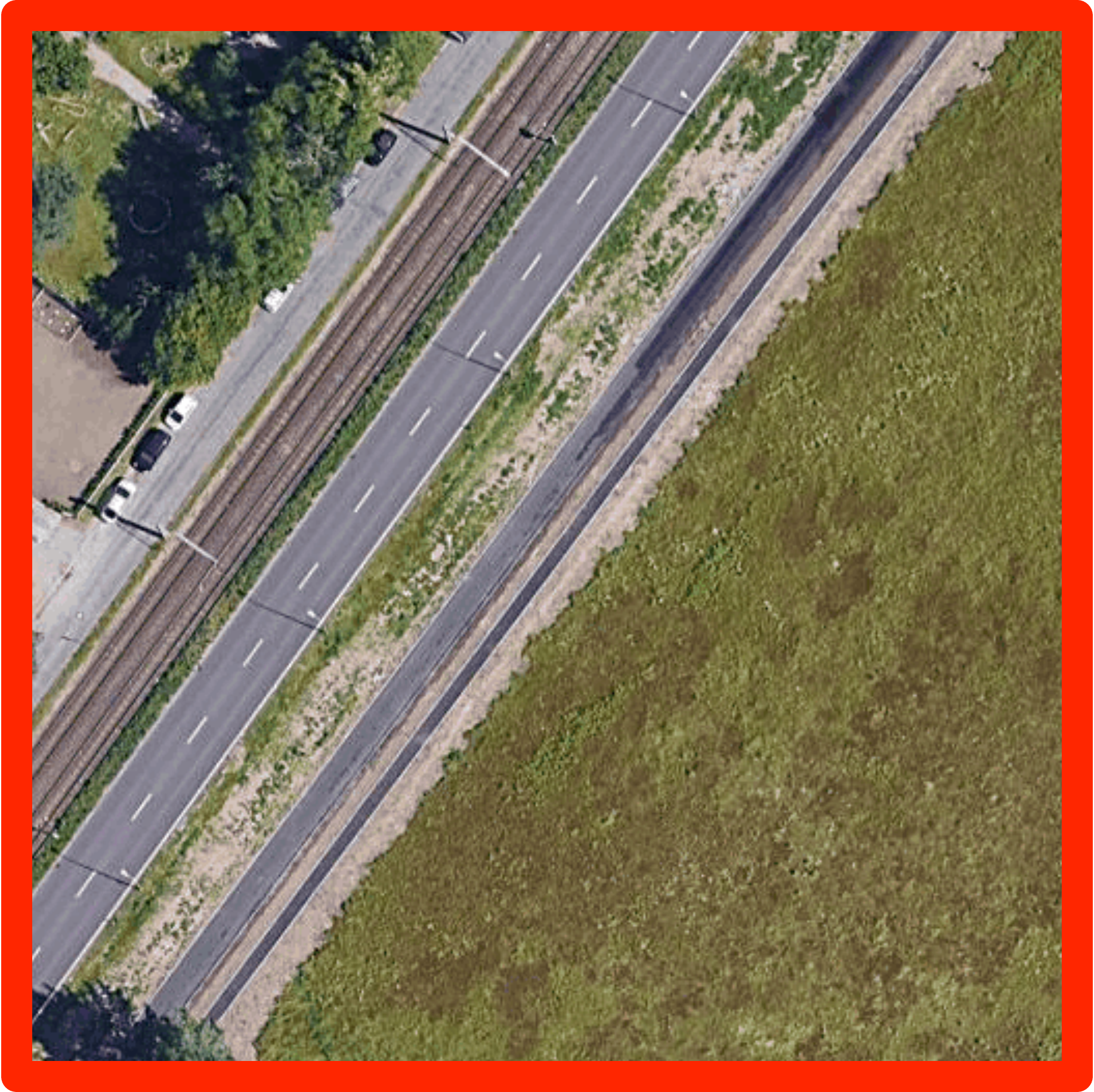}
    \centerline{\scriptsize 9 m}
    \includegraphics[width=\linewidth]{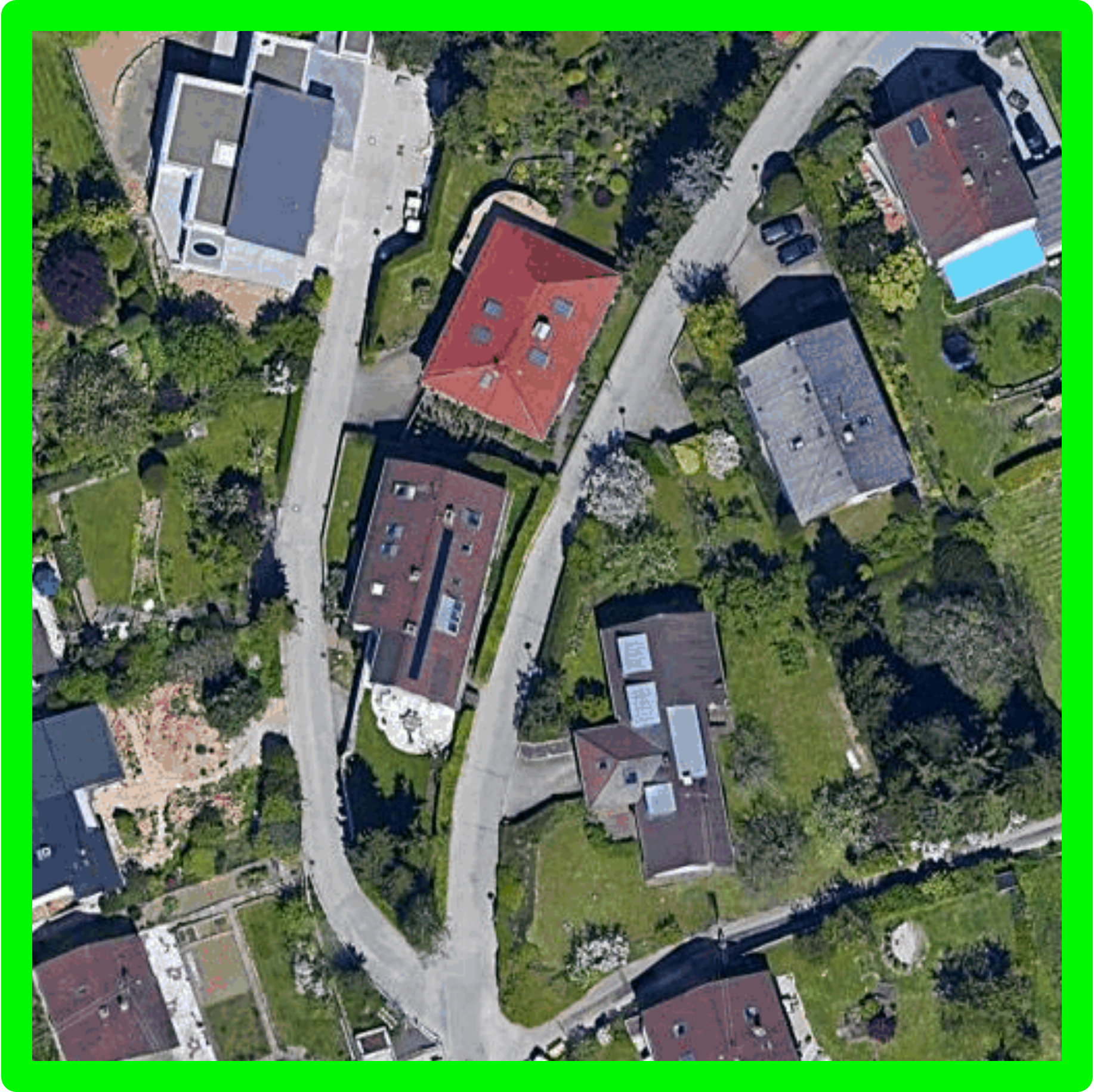}
    \centerline{\scriptsize 7 m}
    \includegraphics[width=\linewidth]{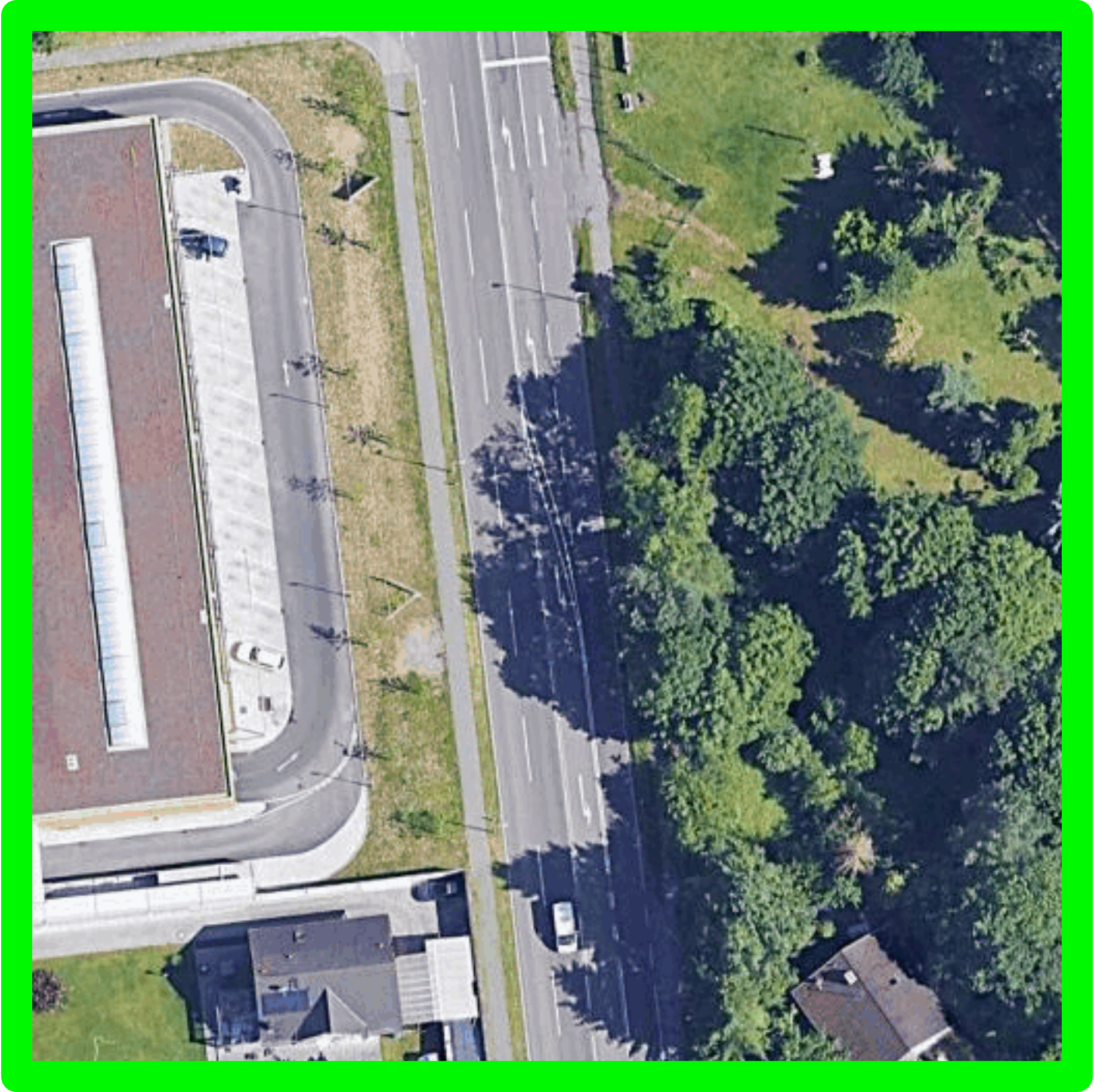}
    \centerline{\scriptsize 5939 m}
    \includegraphics[width=\linewidth]{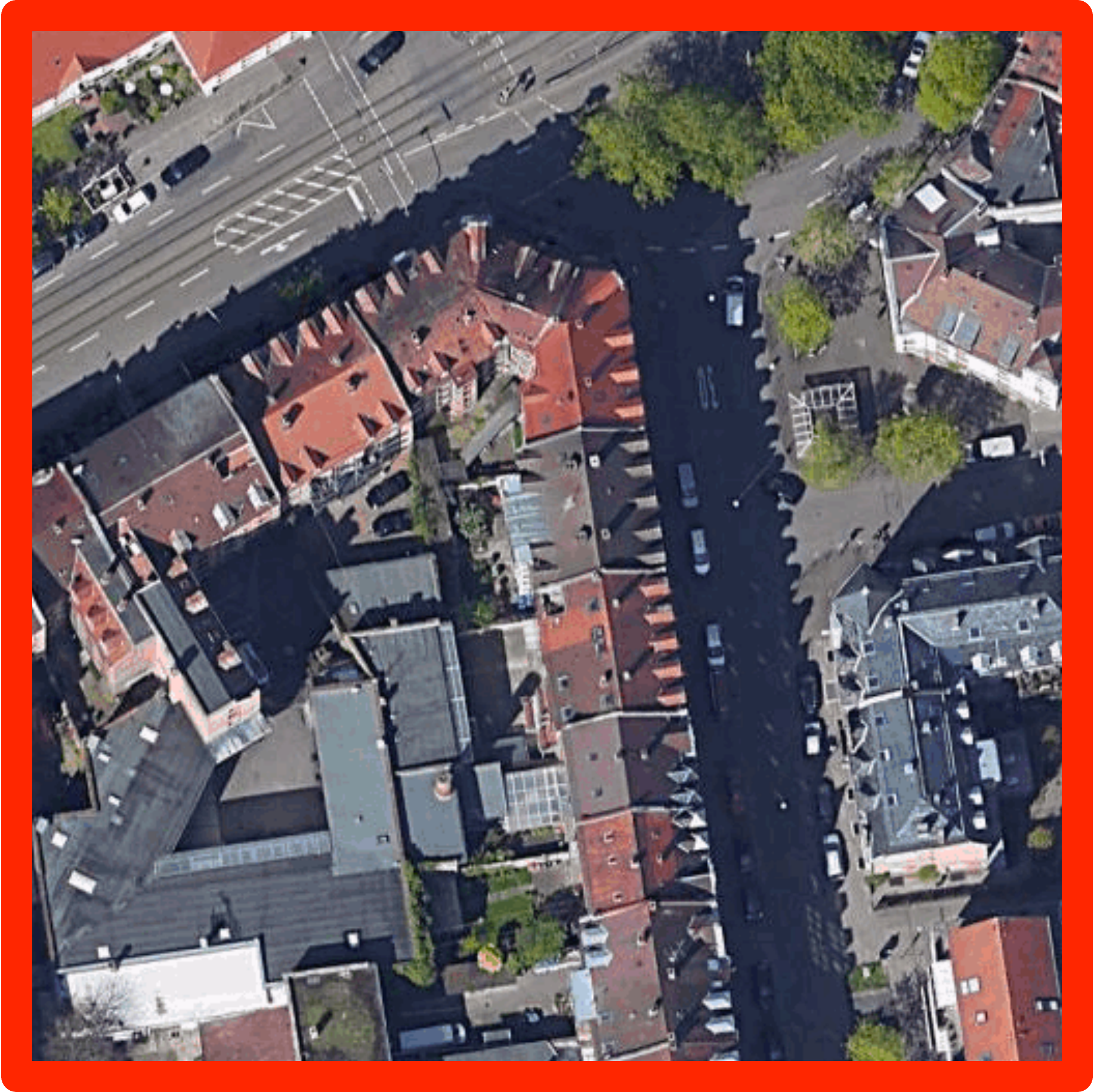}
    \end{minipage}
    }
    \captionof{figure}{\scriptsize Qualitative visualization of retrieved results using 4 image frames in a video. }
    \label{fig:retrieval}
\end{minipage}
\vspace{-1em}
\end{figure}

\subsubsection{Varying sequence lengths.}

One desired property for a video-based localization method is to be robust to various input video lengths after a model is trained. 
Hence, we investigate the performance of our method on different query video sequences (1-16) using a model trained on sequence 4. 
Fig.~\ref{fig:sequence_length} shows that the performance increases elegantly with the increase in number of video sequences. 
This confirms our general intuition that more input images will increase the discriminativeness of the query place and help boost the localization performance. 
Note that when keep increasing the sequence length until the cameras at previous time steps exceed the pre-set coverage of the projected features, the performance will not increase but stay same because we did not fuse exceeded information. 
Scene contents too far from the query camera location are also less useful for localization. 

When only using one image for localization, the feature extraction time for a query descriptor is $0.15$s. 
With the increasing of sequence numbers, the query descriptor extraction time increases linearly. 
 We expect this can be accelerated by parallel computation. 
The retrieval time for each ground image on Test-2 is around 3ms, and the coverage of satellite images in Test-2 is about $710, 708 \ \text{m}^2$.
It takes 8GB GPU memory when the sequence=4 and 24GB when the sequence=16.
We show some qualitative examples of retrieved results in Fig.~\ref{fig:retrieval} using sequence number 4.

\subsection{Single image-based localization}
Single image-based localization is a special case of video-based localization, \ie, when the image frame count in the video is one. 
In this section, we compare the performance of our method with the recent state-of-the-art (SOTA) that are invented for cross-view single image-based localization, including CVM-NET~\cite{Hu_2018_CVPR}, CVFT~\cite{shi2020optimal}, SAFA~\cite{shi2019spatial}, Polar-SAFA~\cite{shi2019spatial}, DSM~\cite{shi2020looking}, Zhu~\etal~\cite{zhu2021vigor}, and Toker~\etal~\cite{toker2021coming}. 
The results are presented in Tab.~\ref{tab:stoa}. 
It can be seen that our method significantly outperforms the recent SOTA algorithms.

Among the compared algorithms, DSM~\cite{shi2020looking} achieves the best performance, because it explicitly addresses the challenge of limited FoV problem of query images while the others assume that query images are full FoV panoramas. 
By comparing SAFA and Polar-SAFA, we can observe that the polar transform boosts the performance on Test-1 (one-to-one matching) while impairs the performance on Test-2 (one-to-many matching). This is consistent with the conclusion in Shi~\etal~\cite{shi2019spatial} and Zhu~\etal~\cite{zhu2021vigor}. 

Based on SAFA, Zhu~\etal~\cite{zhu2021vigor} proposes two training losses: (1) an IoU loss and (2) a GPS loss.  
However, we do not found the two items work well on the KITTI-CVL dataset. 
We guess that the IoU loss is only suitable for the panorama case. 
The limited FoV images in the KITTI-CVL dataset have a smaller overlap with satellite images than panoramas, and thus the original IoU loss may not provide correct guidance for training. 
The GPS loss does not help mainly because of the inaccuracy of the GPS data in our dataset. We provide the GPS accuracy analysis of the KITTI dataset in the supplementary material. In contrast, our method does not rely on the accurate GPS tags of ground or satellite images. 

\begin{table*}[t!]
\setlength{\abovecaptionskip}{0pt}
\setlength{\belowcaptionskip}{0pt}
\setlength{\tabcolsep}{6pt}
\centering
\scriptsize 
\caption{\scriptsize Comparison with the recent state-of-the-art on single image based localization}
\begin{tabular}{c|HHHHcccc|cccc}
\toprule
\multirow{2}{*}{Method} & \multicolumn{4}{H}{Validation}                                            & \multicolumn{4}{c|}{Test-1}                                         & \multicolumn{4}{c}{Test-2}                                         \\
                        & r@1            & r@5            & r@10           & r@100           & r@1            & r@5            & r@10           & r@100          & r@1            & r@5            & r@10           & r@100          \\\midrule
CVM-NET~\cite{Hu_2018_CVPR}                 & 13.34          & 29.93          & 45.55          & 98.14           & 6.43           & 20.74          & 32.47          & 84.07          & 1.01           & 4.33           & 7.52           & 32.88          \\
CVFT~\cite{shi2020optimal}                    & 11.35          & 29.85          & 46.23          & 96.44           & 1.78           & 7.20           & 14.40          & 73.55          & 0.20           & 1.29           & 3.03           & 16.86          \\
SAFA~\cite{shi2019spatial}                    & 43.95          & 73.58          & 87.13          & 100.00          & 4.89           & 15.77          & 23.29          & 87.75          & 1.62           & 4.73           & 7.40           & 30.13          \\
Polar-SAFA~\cite{shi2019spatial}              & 34.55          & 65.79          & 84.21          & 99.87           & 6.67           & 17.06          & 27.62          & 86.53          & 1.13           & 3.76           & 6.23           & 28.22          \\
DSM~\cite{shi2020looking}                     & 23.12          & 36.96          & 43.48          & 74.81           & 13.18          & 41.16          & 58.67          & 97.17          & 5.38           & 18.12          & 28.63          & 75.70          \\
Zhu~\etal~\cite{zhu2021vigor}                   & 19.39          & 41.70          & 56.60          & 95.55           & 5.26           & 17.79          & 28.22          & 88.44          & 0.73           & 3.28           & 5.66           & 27.86          \\
Toker~\etal~\cite{toker2021coming}                  & \textbf{65.68} & \textbf{88.35} & \textbf{95.34} & \textbf{100.00} & 2.79           & 7.72           & 11.69          & 58.92          & 2.39           & 5.50           & 8.90           & 27.05          \\
\textbf{Ours}                    & 42.58          & 66.65          & 79.66          & 98.52           & \textbf{17.71} & \textbf{44.56} & \textbf{62.15} & \textbf{98.38} & \textbf{9.38} & \textbf{24.06} & \textbf{34.45} & \textbf{85.00} \\\bottomrule
\end{tabular}
\label{tab:stoa}
\vspace{-1em}
\end{table*}

\begin{table}[t!]
\setlength{\tabcolsep}{0.8pt}
\centering
\scriptsize
\caption{Comparison results on CVUSA and CVACT (recall at top-1).}
\begin{tabular}{ccccc|ccccc}
\toprule
\multicolumn{5}{c|}{CVUSA}                                                                                                                                                                                                & \multicolumn{5}{c}{CVACT}                                                                                                                                                                                                \\
SAFA\cite{shi2019spatial} & Polar-SAFA\cite{shi2019spatial} & DSM\cite{shi2020looking} & Zhu \etal\cite{zhu2021vigor} & \textbf{Ours}           & SAFA\cite{shi2019spatial} & Polar-SAFA\cite{shi2019spatial} & DSM\cite{shi2020looking} & Zhu \etal\cite{zhu2021vigor} & \textbf{Ours}           \\
68.03                                          & 72.15                                                &       63.17                                        &          53.77                                       & \textbf{72.75} & 56.69                                          & 62.71                                                & 55.07                                         & 49.45                                           & \textbf{66.30}
 \\ \bottomrule
\end{tabular}
\label{tab:USA&ACT}
\vspace{-1em}
\end{table}

\noindent\textbf{CVUSA \& CVACT. } 
Our method applies to panoramas as long as camera parameters are given. 
However, they are not available in {the existing panorama datasets, e.g.,} {\small CVUSA} and {\small CVACT}. 
Furthermore, the matching satellite image centers align precisely with query locations in the two datasets, which is not in practice. 
To make the experiments meaningful, we (i) approximate the camera parameters of the two datasets by visual and geometry verifications; and (ii) randomly translate satellite images (0-36 pixels) to make their centers not aligned with query locations. 
Results are shown in Tab.~\ref{tab:USA&ACT}. It can be seen our method outperforms SOTA methods. 
The SOTA results are inferior to that in their original paper because of the practical setting (as in (ii)).

\subsection{Limitations}
Our method assumes that the north direction is provided by a compass, following previous works~\cite{Liu_2019_CVPR,shi2019spatial,shi2020looking,toker2021coming,zhu2021vigor}, and the absolute scale of camera translations can be estimated roughly from the vehicle velocity. 
We have not investigated how significant tilt and roll angle changes will affect the performance, because the tilt and roll angles in the KITTI dataset are very small and we set them to zero. 
In autonomous driving scenarios, the vehicle-mounted cameras are usually perpendicular to the ground plane. Thus there are only slight changes in tilt and toll during driving.

\section{Conclusions}
This paper introduced a novel geometry-driven semantic correspondence learning approach for cross-view video-based localization. 
Our method includes a Geometry-driven View Projection block to bridge the cross-view domain gap, a Photo-consistency Constrained Sequence Fusion module to aggregate the sequential ground-view observations and a Scene-prior driven similarity matching mechanism to determine the location of a ground camera with respect to a satellite image center.  
Benefiting from the proposed components, we demonstrate that using a video rather than a single image for localization significantly facilitates the localization performance considerably.

\bibliographystyle{splncs}
\bibliography{egbib}


\newpage
\appendix

\section{Dataset Statistics}

\begin{figure}[ht!]
 \setlength{\abovecaptionskip}{0pt}
    \setlength{\belowcaptionskip}{0pt}
    \centering
    \adjincludegraphics[width=\linewidth,trim={{0} {\width/20} {0} {0}},clip]{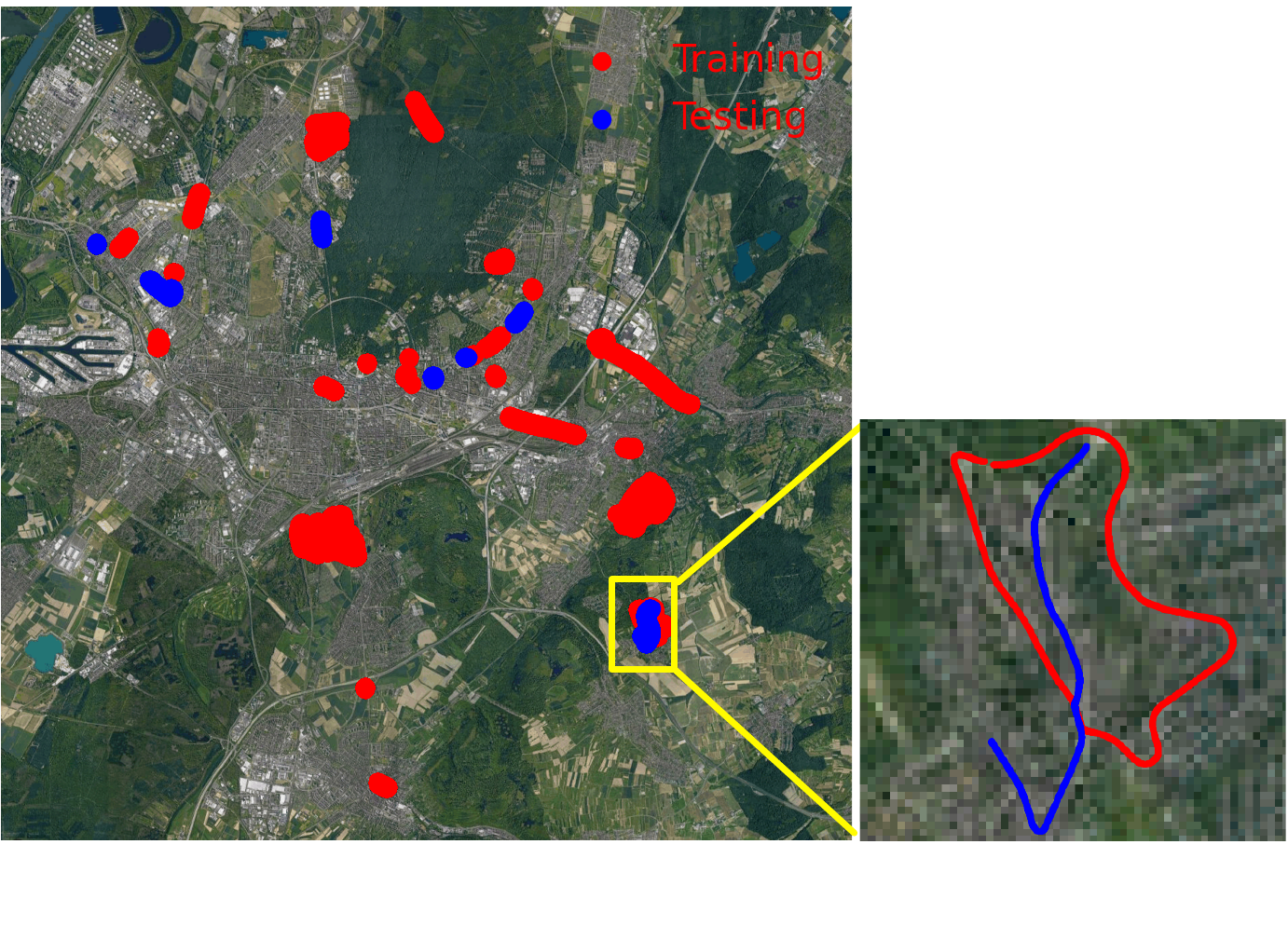}
    \caption{Training and testing data distribution of the introduced KITTI-CVL. They are captured from different regions.}
    \label{fig:data}
\end{figure}

In this section, we provide more illustrations of the introduced KITTI-CVL dataset. 
Fig.~\ref{fig:data} presents an overview of the sampling distributions of the training and testing sets, where training images are captured from the red area, and testing images are sampled in the blue region. 
The training and testing sets do not overlap. 
The Validation set is sampled from the same area of the training set.

\begin{figure}[ht!]
 \setlength{\abovecaptionskip}{2pt}
    \setlength{\belowcaptionskip}{0pt}
    \centering
    \subfigure[Test-1]{
    \centering
    \includegraphics[width=0.45\linewidth]{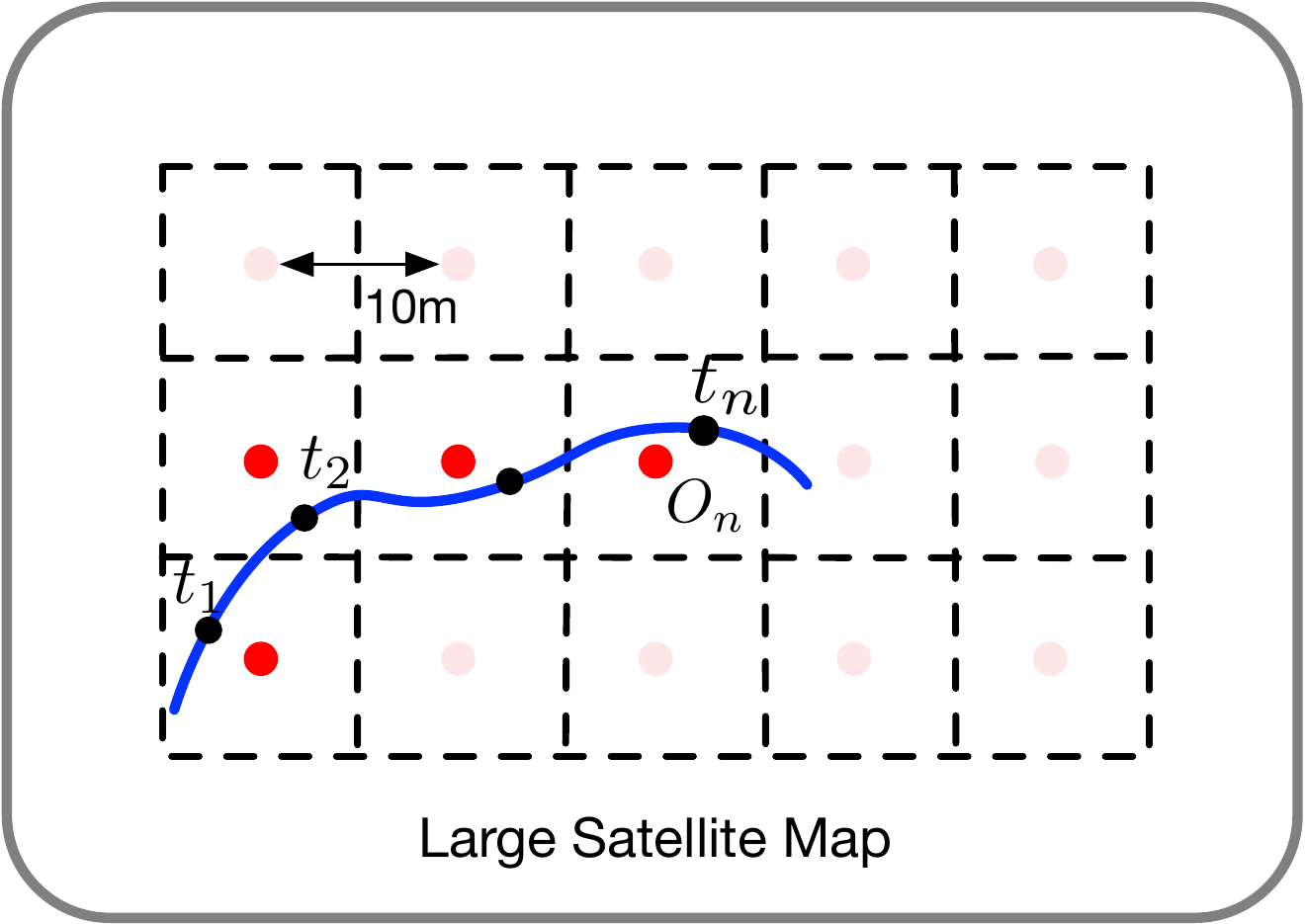}
    \label{subfig:test1}}
    \hspace{0.2em}
    \subfigure[Test-2]{
    \centering
    \includegraphics[width=0.45\linewidth]{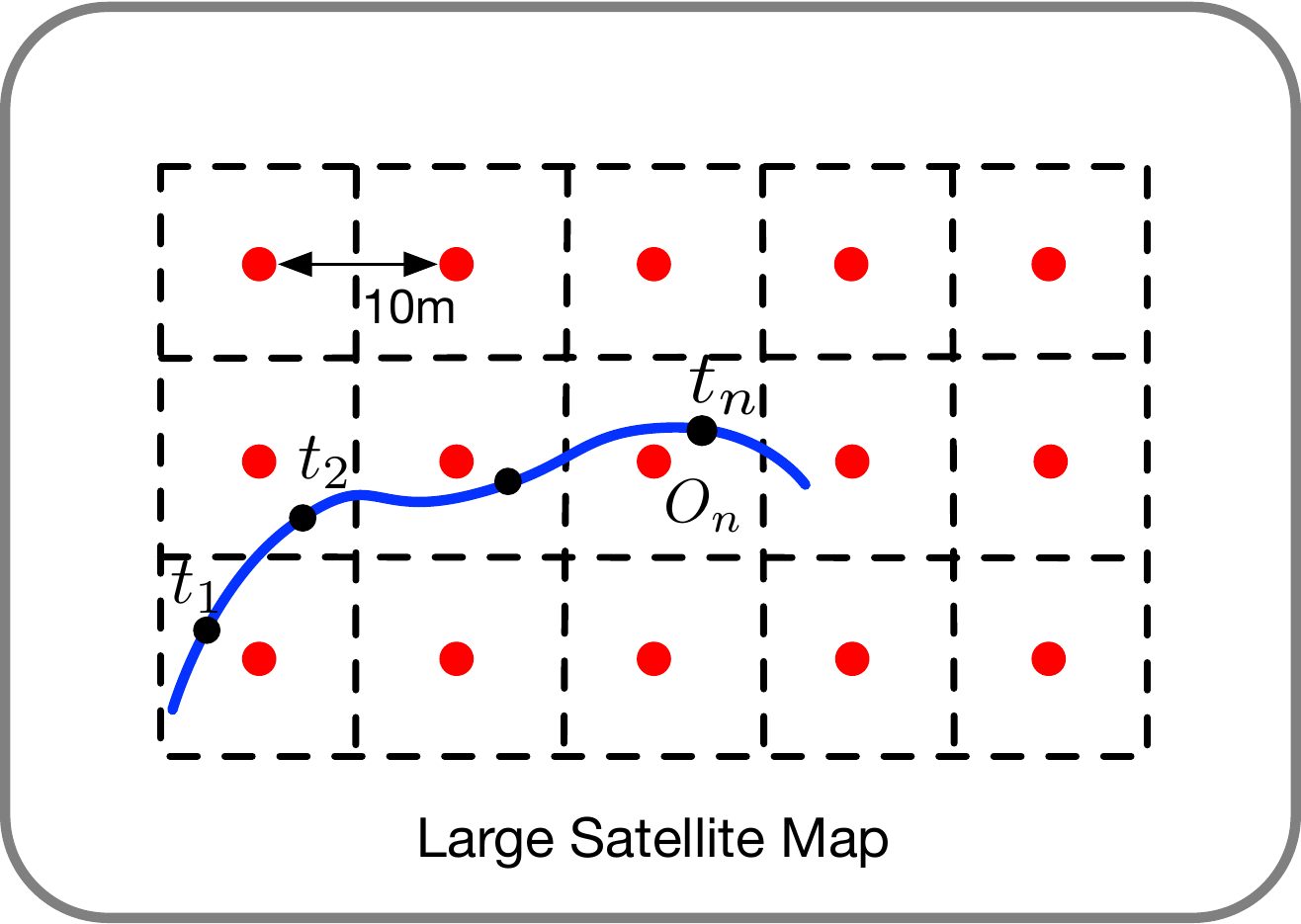}
    \label{subfig:test2}}
    \caption{ Comparison between (a) Test-1 and (b) Test-2. 
    Here, the large black box indicates the whole region of interest. The red dots represent the satellite image centers in the database. They are sampled every ten meters in the region of interest. The blue curve represents a trajectory of a camera. The black dots on the curve denote the camera locations at different time steps.
    In Test-1, only the nearest satellite image for each query ground image is retained in the database. Thus, we mask the other satellite image centers in the grid, {which can be regarded as a non-distractor case.}
    In Test-2, all the satellite images sampled in the grid are reserved.
    }
    \label{fig:Test1_Test2}
\end{figure}

\begin{figure}[]
 \setlength{\abovecaptionskip}{5pt}
    \setlength{\belowcaptionskip}{0pt}
    \centering
    \begin{minipage}{0.32\linewidth}
    \includegraphics[width=\linewidth, height=0.45\linewidth]{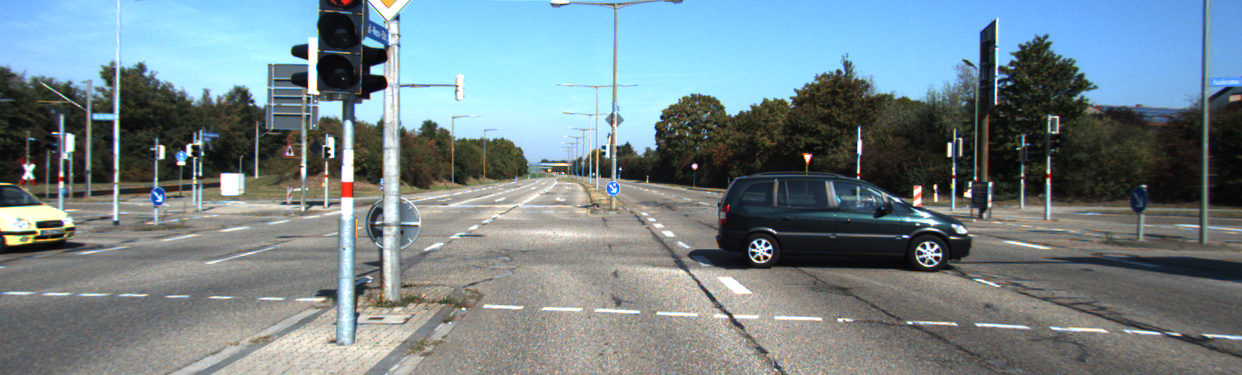}
    \centerline{\scriptsize lat = 49.025527{\color{red}549337}}
    \centerline{\scriptsize lon =  8.448562{\color{red}3653485}}
    \end{minipage}
    \begin{minipage}{0.32\linewidth}
    \includegraphics[width=\linewidth, height=0.45\linewidth]{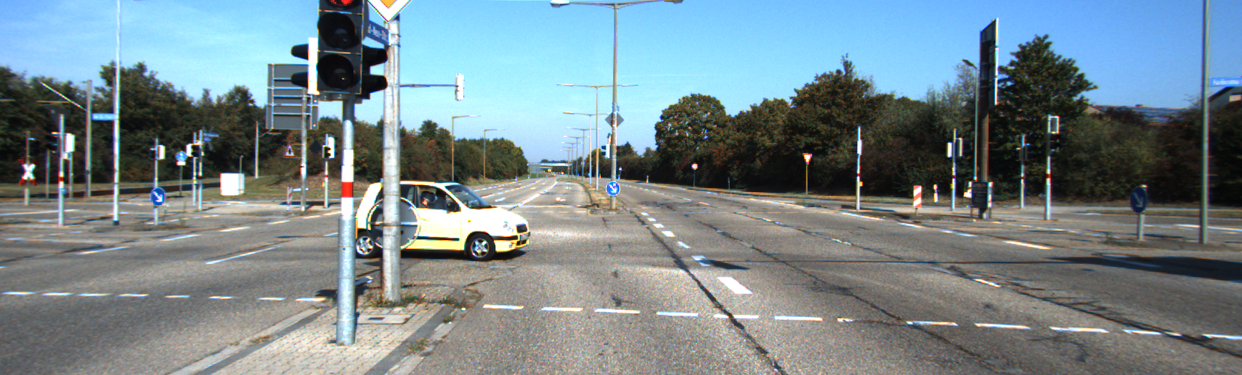}
    \centerline{\scriptsize lat = 49.025527{\color{red}483717}}
    \centerline{\scriptsize  lon = 8.448562{\color{red}3463895}}
    \end{minipage}
    \begin{minipage}{0.32\linewidth}
    \includegraphics[width=\linewidth, height=0.45\linewidth]{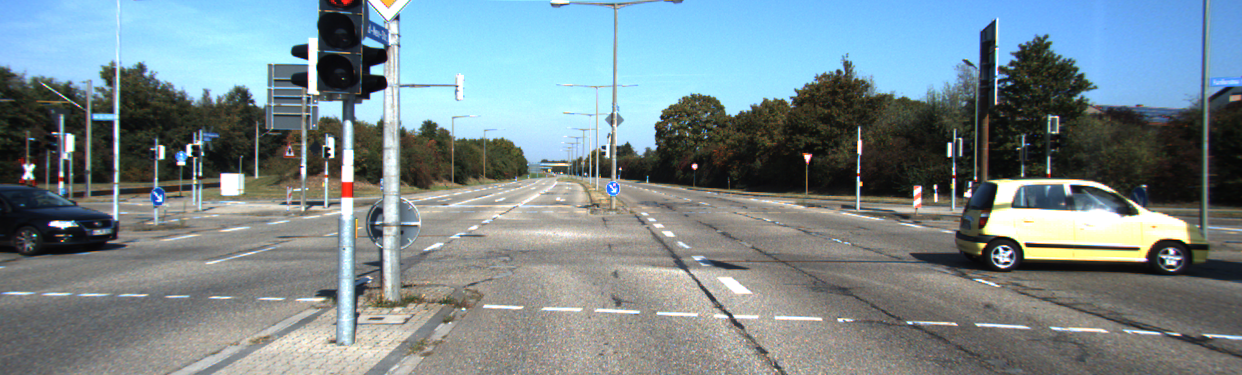}
    \centerline{\scriptsize lat = 49.025527{\color{red}419798}}
    \centerline{\scriptsize lon =  8.448562{\color{red}3793197}}
    \end{minipage}
    \caption{\scriptsize Varying GPS tags for a static camera. The distances between the tags are from $0.01$m to $0.3$m. 
    Images are from drive ``2011\_09\_26\_drive\_0017\_sync''.
    }
    \label{fig:gps_not_moving}
\end{figure}

\begin{figure}[]
 \setlength{\abovecaptionskip}{2pt}
    \setlength{\belowcaptionskip}{0pt}
    \centering
    \begin{minipage}{0.5\linewidth}
    \includegraphics[width=\linewidth, height=0.4\linewidth]{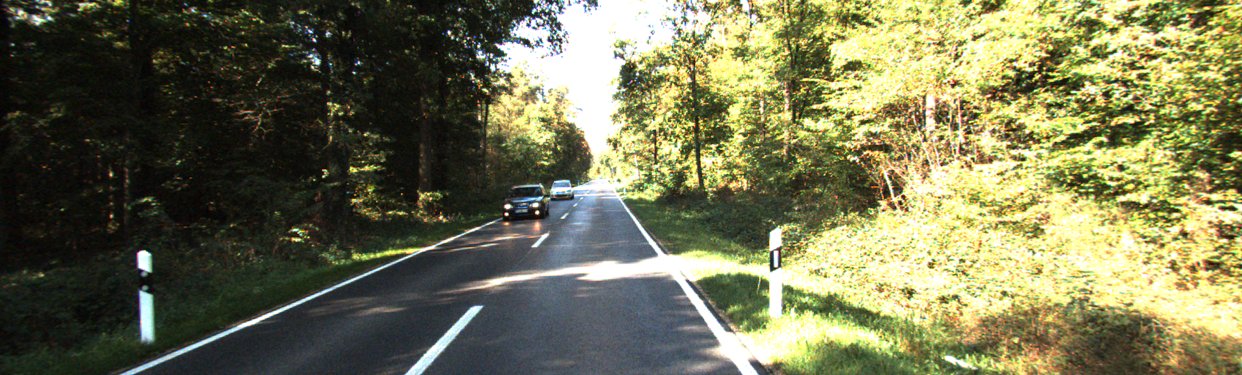}
    \centerline{\scriptsize Ground Image}
    \end{minipage}
    \hspace{4em}
    \begin{minipage}{0.21\linewidth}
    \includegraphics[width=\linewidth]{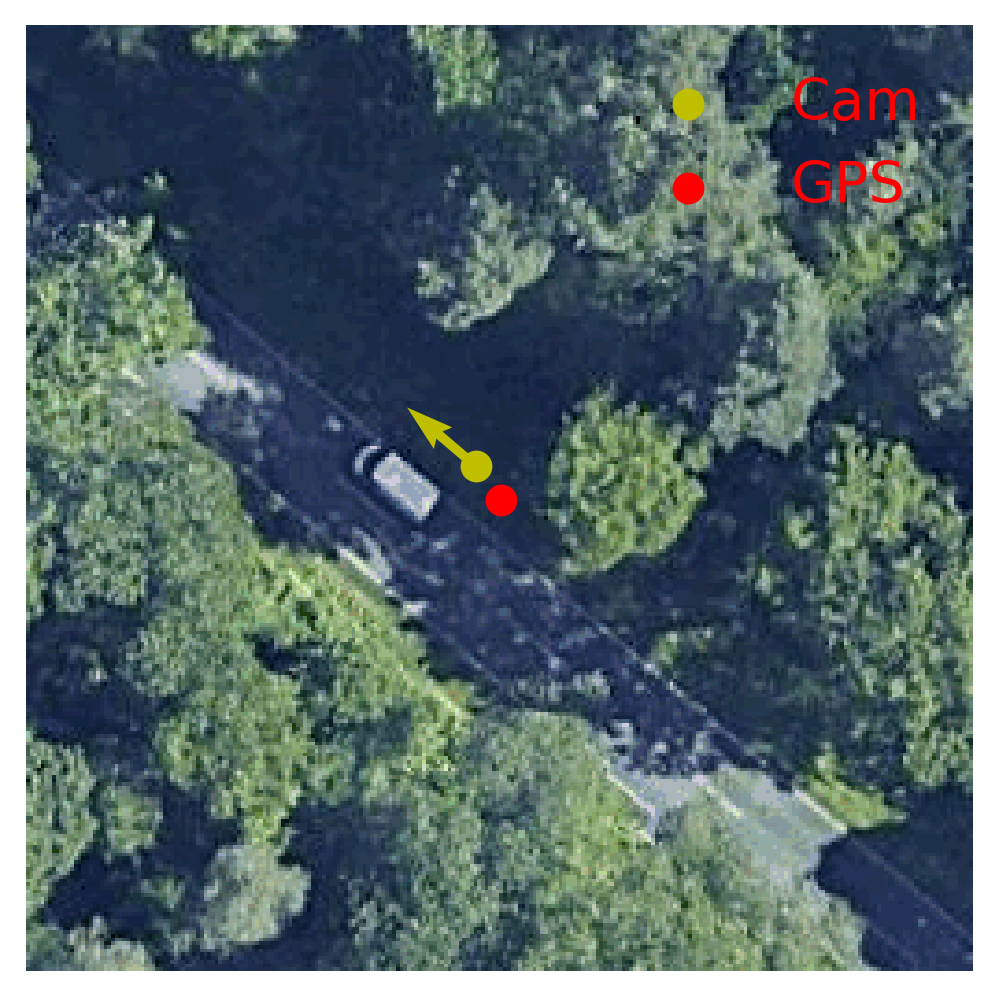}
    \centerline{\scriptsize Satellite Image}
    \end{minipage}
    \caption{\scriptsize The query ground image is captured on the road, while its position from GPS data shows it is on top of a vegetation area. The error is around $1$m. In the satellite image, the red point indicates the position of the GPS device on the car of the KITTI dataset, and the yellow point indicates the position of the left color camera. The yellow arrow tags the vehicle heading direction. 
    Images are from drive ``2011\_09\_26\_drive\_0027\_sync''.
    }
    \label{fig:gps}
\end{figure}

\begin{figure}[t!]
\setlength{\abovecaptionskip}{2pt}
    \centering
    \begin{minipage}{0.48\linewidth}
    \includegraphics[width=\linewidth, height=0.45\linewidth]{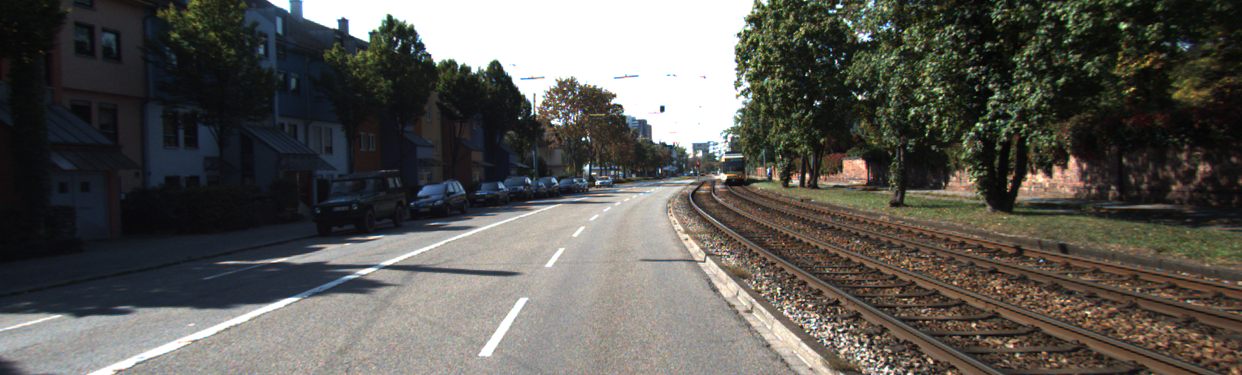}
    \includegraphics[width=\linewidth, height=0.45\linewidth]{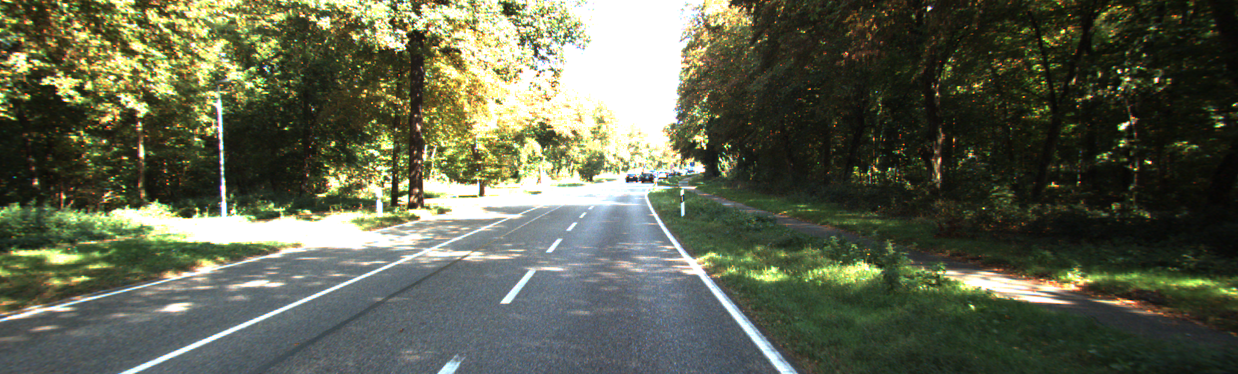}
    \centerline{\scriptsize Ground Image}
    \end{minipage}
    \begin{minipage}{0.22\linewidth}
    \includegraphics[width=\linewidth]{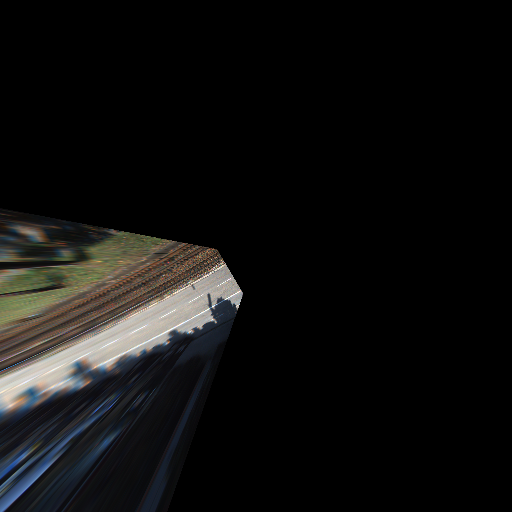}
    \includegraphics[width=\linewidth]{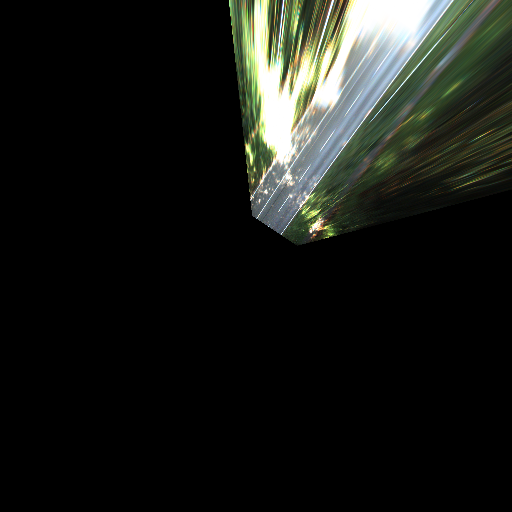}
    \centerline{\scriptsize Projected Image}
    \end{minipage}
    \begin{minipage}{0.22\linewidth}
    \adjincludegraphics[width=\linewidth,trim={{\width/4} {\width/4} {\width/4} {\width/4}},clip]{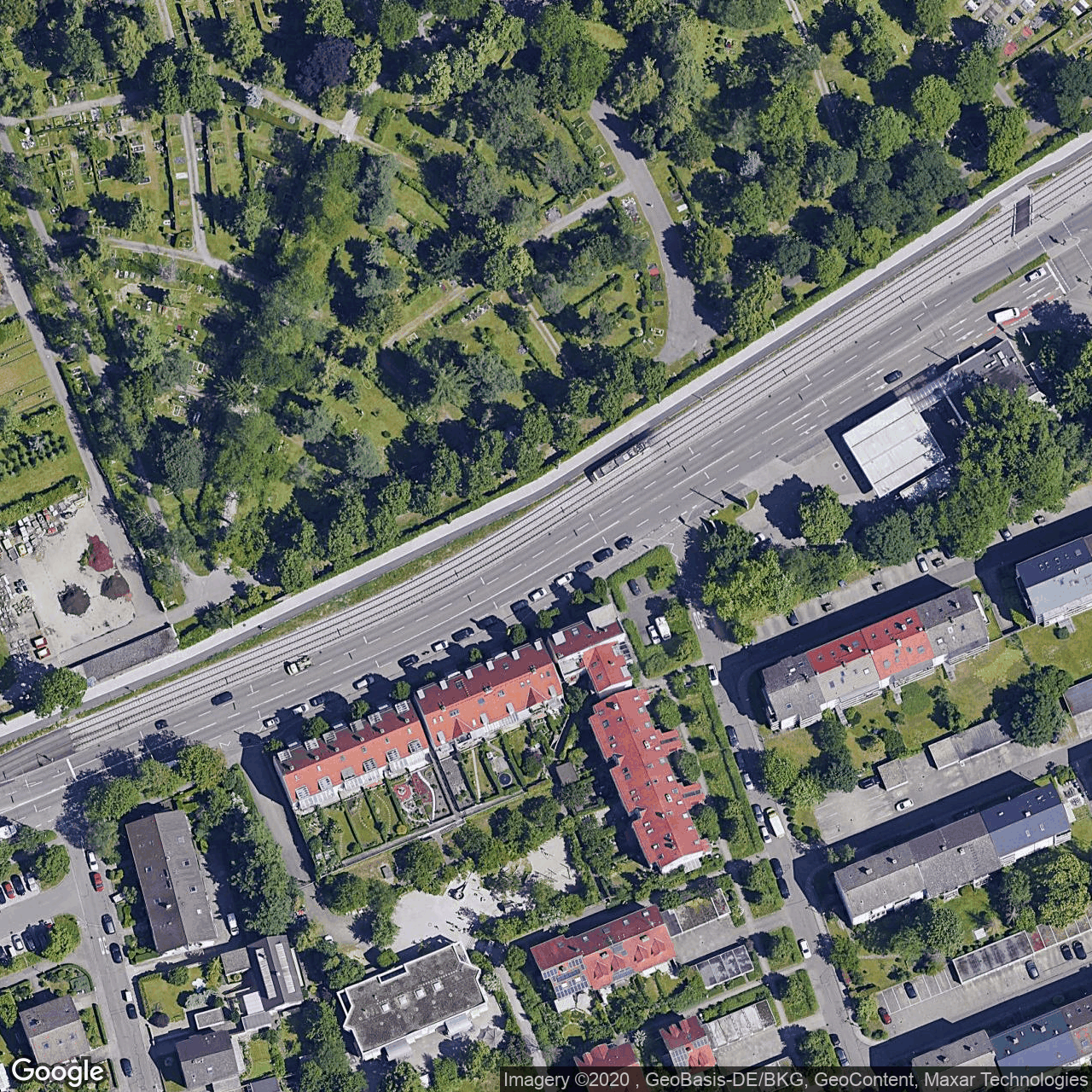}
    \adjincludegraphics[width=\linewidth,trim={{\width/4} {\width/4} {\width/4} {\width/4}},clip]{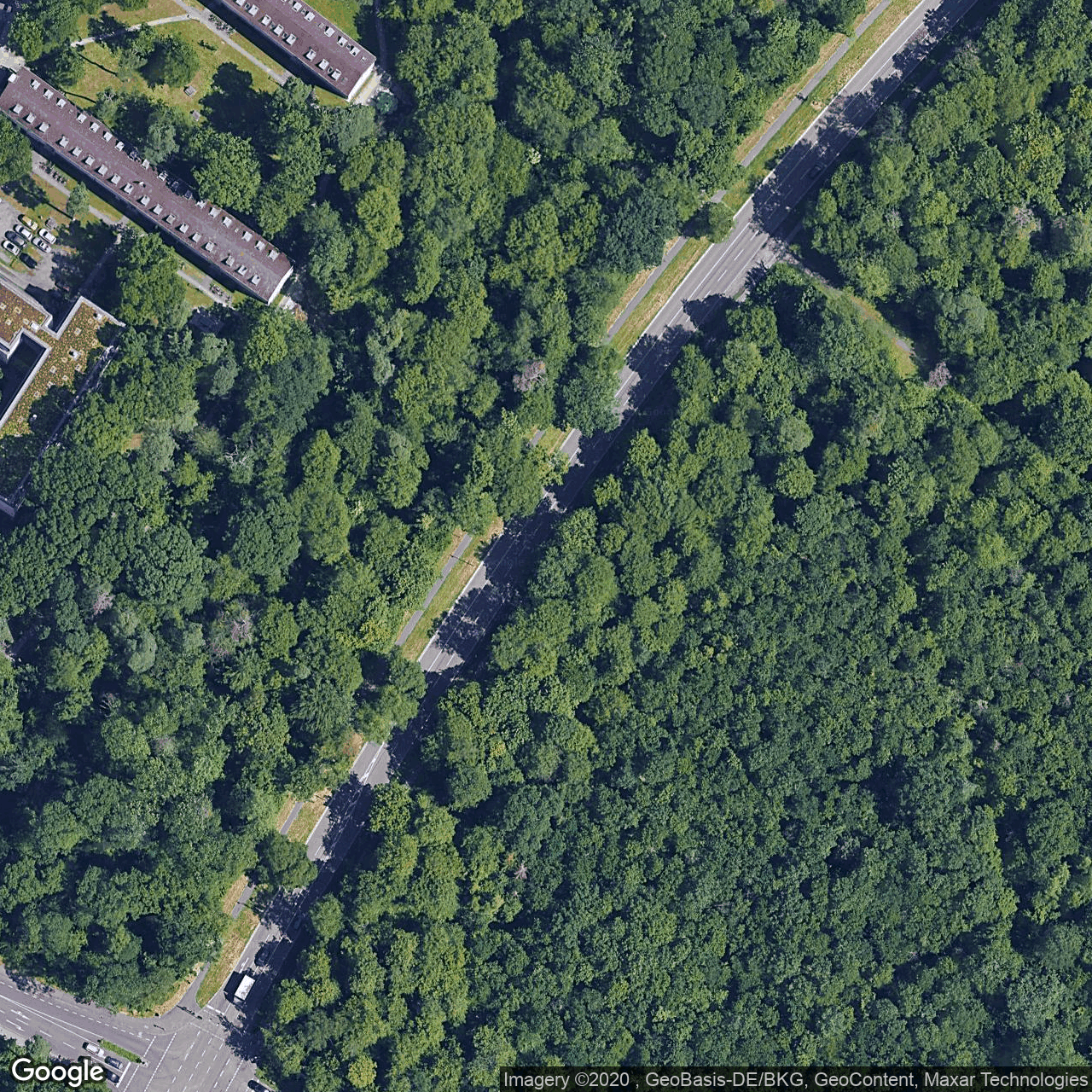}
    \centerline{\scriptsize Satellite Image}
    \end{minipage}
    \caption{\scriptsize Visualization of projected images in the overhead-view.}
    \label{fig:proj_img}
\end{figure}

\textbf{Differences between Test-1 and Test-2.}
Our two test sets, \ie, Test-1 and Test-2, share the same query ground images. 
Their differences lie in satellite images in the database.
As shown in the left of Fig.~\ref{fig:Test1_Test2}, in Test-1, only the nearest satellite image of each query image is retained in the database. While in Test-2, all the satellite images within the large area are reserved. 
Test-2 has more distracting images in the database and thus is a more challenging test set than Test-1.

\smallskip 
\textbf{GPS noise.} 
We found there is slight noise in the GPS raw data provided by KITTI. 
For example, as shown in Fig.~\ref{fig:gps}, the ground camera is not moving, while the provided GPS data varies for these images. 
Fig.~\ref{fig:gps} presents another example where the camera is expected to be on the road while the camera location provided by the GPS data is on top of the vegetation area. 
We guess those minor errors make the GPS loss proposed in Zhu~\etal~[51] not work well in the KITTI-CVL dataset.

\begin{table*}[t!]
\setlength{\abovecaptionskip}{0pt}
\setlength{\belowcaptionskip}{0pt}
\setlength{\tabcolsep}{4pt}
\centering
\scriptsize
\caption{\scriptsize Additional ablation study results of our method}
\begin{tabular}{c|HHHHcccc|cccc}
\toprule
\multirow{2}{*}{Method} & \multicolumn{4}{H}{Validation}                                            & \multicolumn{4}{c|}{Test-1}                                         & \multicolumn{4}{c}{Test-2}                                         \\
                        & r@1            & r@5            & r@10           & r@100           & r@1            & r@5            & r@10           & r@100          & r@1            & r@5            & r@10           & r@100          \\\midrule

Ours (GVP on Img)                   & 4.41	&11.65	&19.28	& 70.89	& 1.33	& 5.82	& 10.23	& 54.14	& 0.16	& 0.36	& 0.93	& 13.18          \\
Ours w/o High Objects                 & \textbf{43.98} & \textbf{68.73} & \textbf{83.86} & \textbf{99.28} & 2.35	&6.67	&10.76	&52.45	&0.65	&2.71	&3.80	&23.49         \\
\textbf{Ours}  & { 42.58}    & { 66.65}    & { 79.66}    & {98.52} & \textbf{21.80} & \textbf{47.92} & \textbf{ 64.94}    & \textbf{99.07} & \textbf{12.90} & \textbf{27.34} & \textbf{38.62} & \textbf{ 85.00}    \\ \bottomrule
\end{tabular}
\label{tab:addtional}
\end{table*}

\section{Additional Illustrations and Experiments}
To better illustrate that GVP is able to project ground-view images following the geometric constraints, we apply the GVP to the original ground-view images, as shown in Fig.~\ref{fig:proj_img}. 
It can be seen that pixels on the ground plane have been successfully restored in the overhead view. While for pixels above the ground plane, they undergo obvious distortions. Hence, we apply the GVP to feature level rather than image level in our framework. 
The high-level features are expected to establish semantic correspondences between the two views and circumvent the side effects of distortions in projection. 
We provide the performance of applying GVP to image level in the first row of Tab.~\ref{tab:addtional}, denoted as ``Ours (GVP on Img)''. 
Not surprisingly, it is significantly inferior to Ours, where the GVP is applied to the feature level.

\begin{figure}[t!]
 \setlength{\abovecaptionskip}{5pt}
    \centering
    \begin{minipage}{0.48\linewidth}
    \includegraphics[width=\linewidth, height=0.45\linewidth]{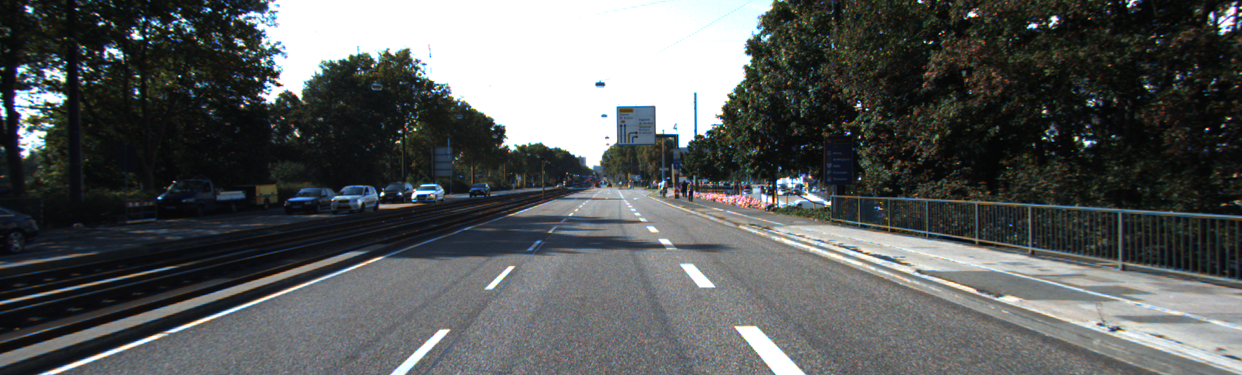}
    \includegraphics[width=\linewidth, height=0.45\linewidth]{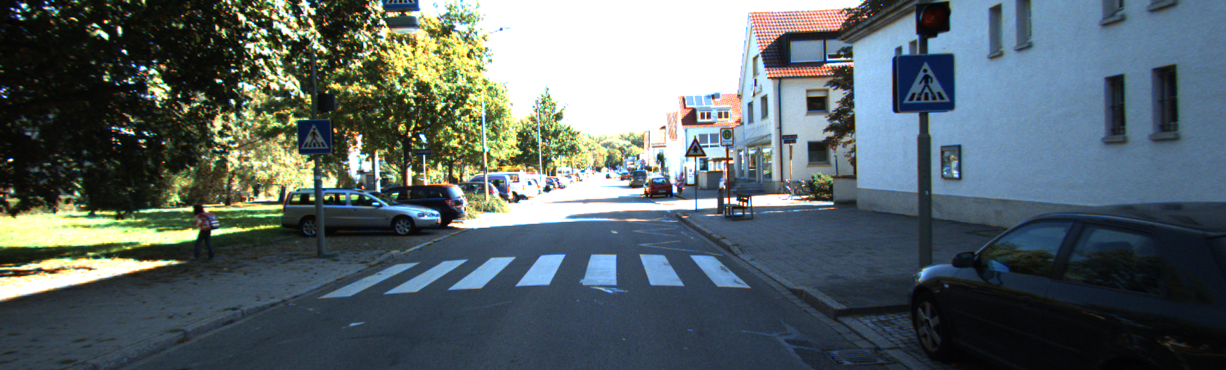}
    \end{minipage}
    \begin{minipage}{0.48\linewidth}
    \includegraphics[width=\linewidth, height=0.45\linewidth]{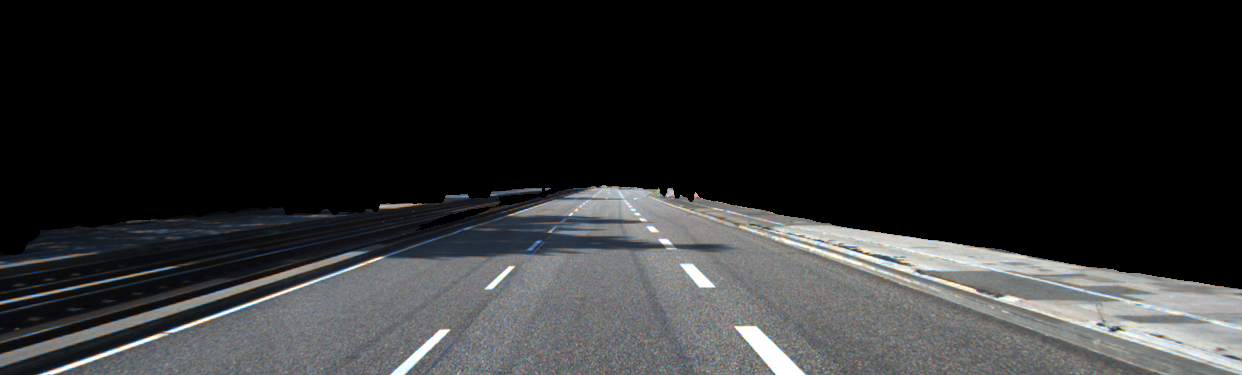}
    \includegraphics[width=\linewidth, height=0.45\linewidth]{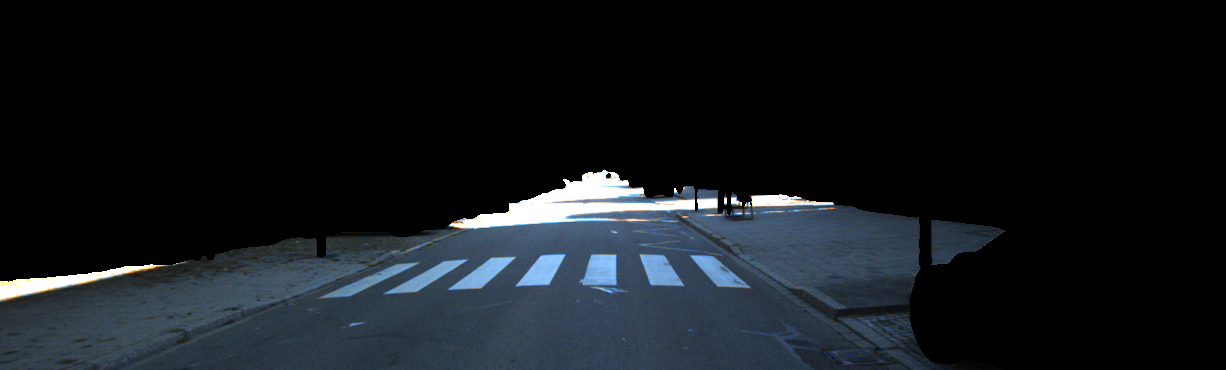}
    \end{minipage}
    \caption{Comparison between original ground images (left) and semantic-filtered ground images (right).}
    \label{fig:seman}
\end{figure}

Next, we use semantic maps to filter out the scene objects above the ground plane, \eg, buildings, trees, sky, etc. from the input ground-view images, and feed the processed images to our network, denoted as ``Ours w/o High Objects''. 
Fig.~\ref{fig:seman} shows the comparison between the original ground-view images and the processed images. 
The performance of ``Ours w/o High Objects'' is presented in the second row of Tab.~\ref{tab:addtional}. 
It can be seen that the performance is also worse than ours, indicating that our GVP module on feature level successfully encodes information of scene objects that have higher heights.

\begin{table}[t!]
\setlength{\tabcolsep}{4pt}
\centering
\scriptsize
\caption{Adding baseline results (sequence=1) for alternative sequence fusion algorithms in Sec. 5.1.2.
}
\begin{tabular}{c|c|c|cccc|cccc}
\toprule
                                                                                   &                         & Seq & \multicolumn{4}{c|}{Test-1}                                        & \multicolumn{4}{c}{Test-2}                                        \\ \midrule
\multirow{4}{*}{Direct Fusion}                                                     & \multirow{1}{*}{Conv2D} & 1   & {1.98}  & 5.30           & 9.50           & 60.01          & 7.60           & 18.48          & 26.49          & 72.71          \\
                                                                                  &                         & 4   & 1.25           & \textbf{5.90}  & \textbf{10.80} & \textbf{65.91} & \textbf{8.90}  & 18.44          & \textbf{26.61} & \textbf{76.51} \\\cmidrule{2-11}
                                                                                   & \multirow{1}{*}{LSTM}   & 1   & 7.44           & 22.16          & 35.06          & 93.97          & 3.36           & 9.87           & 16.17          & 64.46          \\ 
                                                                                  &                         & 4   & \textbf{12.53} & \textbf{32.11} & \textbf{50.42} & \textbf{96.93} & \textbf{5.78}  & \textbf{15.89} & \textbf{23.01} & \textbf{70.60} \\\midrule
\multirow{6}{*}{\begin{tabular}[c]{@{}c@{}}Attention-based \\ Fusion\end{tabular}} & \multirow{1}{*}{Conv2D} & 1   & 15.37          & 41.16          & 56.98          & 95.79          & 9.18           & 20.58          & 30.53          & 76.83          \\
                                                                                  &                         & 4   & \textbf{18.80} & \textbf{47.03} & \textbf{61.75} & \textbf{96.64} & \textbf{11.69} & \textbf{25.03} & \textbf{36.55} & \textbf{81.52} \\\cmidrule{2-11}
                                                                                   & \multirow{1}{*}{LSTM}   & 1   & 11.69          & 37.28          & 56.65          & 96.44          & 7.04           & 18.16          & 26.57          & 79.09          \\
                                                                                  &                         & 4   & \textbf{15.93} & \textbf{47.88} & \textbf{66.03} & \textbf{97.61} & \textbf{9.70}  & \textbf{24.30} & \textbf{35.26} & \textbf{85.08} \\\cmidrule{2-11}
                                                                                   & \multirow{1}{*}{Ours}   & 1   & 17.71          & 44.56          & 62.15          & 98.38          & 9.38           & 24.06          & 34.45          & 78.37          \\
                                                                                  &                         & 4   & \textbf{21.80} & \textbf{47.92} & \textbf{64.94} & \textbf{99.07} & \textbf{12.90} & \textbf{27.34} & \textbf{38.62} & 85.00          \\
\bottomrule
\end{tabular}
\label{tab:single}
\end{table}

\begin{table}[t!]
\setlength{\tabcolsep}{6pt}
\centering
\scriptsize
\caption{Comparison results of our method with or without satellite image height estimation (sequence = 4).
}
\begin{tabular}{cc|cccc|cccc}
\toprule
                                                                                   &                        & \multicolumn{4}{c|}{Test-1}                                        & \multicolumn{4}{c}{Test-2}                                        \\ \midrule
  \multicolumn{2}{c|}{Ours w/ height estimation}                                                                  & 21.57          & 47.41          & 65.16          & 98.30          & 12.15          & 26.70          & 38.72          & 84.95         
\\
\multicolumn{2}{c|}{Ours w/o height estimation}                                                               & {21.80} & {47.92} & {64.94} & {99.07} & {12.90} & {27.34} & {38.62} & 85.00          \\
\bottomrule
\end{tabular}
\label{tab:height}
\end{table}

We also add the baseline results with a sequence number as one for all the sequence fusion alternatives illustrated in Sec. 5.1.2 in the main paper, as shown in Tab.~\ref{tab:single}. Note that ``Conv3D'' is the same as ``Conv2D'' when sequence=1. The results indicate that using a longer sequence generally helps achieve better performance. 
Furthermore, we tried to borrow ideas from MVS to estimate overhead-view satellite image height maps from sequential ground view images. The comparison results between with and without height estimation are shown in Tab.~\ref{tab:height}. There are no significant differences in the final results, but height estimation introduces more computation and memory. Hence, we do not include height estimation in our final method but use the ground-plane assumption in the projection.

\end{document}